\documentclass[11pt]{article}

\usepackage[preprint]{acl}

\usepackage{times}
\usepackage{latexsym}

\usepackage[T1]{fontenc}

\usepackage[utf8]{inputenc}

\usepackage{microtype}

\usepackage{inconsolata}

\usepackage{graphicx}
\usepackage{xcolor}
\usepackage{amssymb}
\usepackage{amsmath}
\usepackage{amsfonts}
\usepackage{enumerate}
\usepackage{booktabs}
\usepackage{multirow}
\usepackage{longtable}
\usepackage{array}
\usepackage{tabularx}

\usepackage[table]{xcolor} 

\definecolor{HeaderGray}{gray}{0.93}

\definecolor{BaselineRow}{RGB}{227,239,255}   
\definecolor{ControlRow}{RGB}{255,242,223}  

\definecolor{BestGreen}{RGB}{228,245,228}

\definecolor{DeltaPosL}{RGB}{235,248,235}   
\definecolor{DeltaPosM}{RGB}{210,240,210}   
\definecolor{DeltaPosH}{RGB}{180,230,180}   

\definecolor{DeltaNegL}{RGB}{255,240,240}   
\definecolor{DeltaNegM}{RGB}{255,220,220}   
\definecolor{DeltaNegH}{RGB}{255,195,195}   

\newcommand{\best}[1]{\cellcolor{BestGreen}\textbf{#1}}
\newcommand{\head}[1]{\cellcolor{HeaderGray}\textbf{#1}}

\title{GeoLAN: Geometric Learning of Latent Explanatory Directions in Large Language Models}

\author{Tianyu Bell Pan \quad \quad Damon L. Woodard \\
  Department of Electrical and Computer Engineering \\
  Florida Institute for National Security \\
  Applied Artificial Intelligence Group \\
  University of Florida \\
  \texttt{tpan1@ufl.edu}, \quad \texttt{dwoodard@ece.ufl.edu}\\}

\begin{document}
\maketitle
\begin{abstract}
Large language models (LLMs) demonstrate strong performance, but they often lack transparency. We introduce GeoLAN, a training framework that treats token representations as geometric trajectories and applies stickiness conditions inspired by recent developments related to the Kakeya Conjecture. We have developed two differentiable regularizers, Katz-Tao Convex Wolff (KT-CW) and Katz-Tao Attention (KT-Attn), that promote isotropy and encourage diverse attention. Our experiments with Gemma-3 (1B, 4B, 12B) and Llama-3-8B show that GeoLAN frequently maintains task accuracy while improving geometric metrics and reducing certain fairness biases. These benefits are most significant in mid-sized models. Our findings reveal scale-dependent trade-offs between geometric precision and performance, suggesting that geometry-aware training is a promising approach to enhance mechanistic interpretability.
\end{abstract}

\vspace{-2mm}
\section{Introduction}\label{sec:label}
\vspace{-2mm}
The deployment of large language models (LLMs) in critical domains, such as legal adjudication, medical diagnostics, and autonomous code generation, has precipitated an urgent interpretability crisis. While these models perform exceptionally well on benchmarks such as MMLU~\citep{hendrycks2020measuring} and GSM8K~\citep{cobbe2021training}, their internal logic operates as a black box, obscuring the causal relationships between inputs and outputs~\citep{lyu2024towards}. This lack of transparency is not just an academic issue; it poses systemic risks regarding inspection, accountability, and safety~\citep{zhang2025sr,orgad2024llms}. When a model fabricates a legal precedent or misdiagnoses a patient, users currently lack the means to audit the reasons behind these errors and often have to rely on the model's potentially fabricated explanations~\citep{guidotti2018survey}. 

A key challenge in tackling this crisis is the inherent limitations of existing interpretability methods. Techniques such as attention visualization, gradient-based saliency maps, and causal mediation analysis provide post hoc explanations~\citep{madsen2022post,chefer2021transformer}. These methods attach plausible human rationales to model outputs without ensuring fidelity to the model's internal mechanisms. More advanced mechanistic interpretability methods, such as sparse autoencoders (SAEs) and dictionary learning, aim to decompose representations into monosemantic features~\citep{nanda2023progress,rai2024practical,cunningham2023sparse,tang2023explainable}. However, these methods face a challenge known as superposition, in which neural networks can represent more features than they have neurons by encoding them in non-orthogonal directions~\citep{klindt2025superposition}. This phenomenon results in individual neurons becoming polysemantic, responding to multiple concepts (e.g., a neuron may activate to both ``cats'' and ``biblical verses''), complicating efforts to achieve clear decomposition. 

In addition to these interpretability challenges, there is the issue of representation degradation, particularly anisotropy~\citep{klindt2025superposition,machina2024anisotropy}. Research has consistently demonstrated that the embedding spaces of standard Transformers are highly anisotropic, meaning that most of the variance is concentrated in a few dominant directions, effectively collapsing the representation into a narrow cone~\citep{razzhigaev2024shape}. This geometric collapse has two harmful effects: it limits the model's effective capacity by underutilizing available dimensional space, and it worsens entanglement, forcing distinct semantic concepts to share a limited subspace. Recent studies using metrics such as IsoScore have confirmed that this anisotropy correlates with a diminished downstream performance and model brittleness~\citep{rudman2022isoscore}. Although some architectures, such as Pythia~\citep{biderman2023pythia}, have shown improved isotropy, the issue persists across most optimized causal language models. More related work is presented in Appendix~\ref{appx:related_work}.

To address these combined challenges, we propose GeoLAN, a framework that treats interpretability as a geometric constraint to be enforced during training rather than discovered post hoc. We hypothesize that preventing representation collapse and compelling the model to use its full geometric space through regularization can yield a disentangled latent structure. Our approach draws inspiration from the Kakeya needle problem in geometric measure theory, particularly the recent resolution of the Kakeya conjecture in $\mathbb{R}^3$, which introduced the concept of sticky Kakeya sets~\citep{wang2025volume}. By viewing token representations not as static points but as continuous trajectories or tubes moving through the network layers, we adapt the mathematical framework of Kakeya sets to the linguistic domain. 

The following \textbf{\underline{core questions}} guide our research: 
\textbf{RQ1:} Can we mathematically formalize the ``messiness'' of latent spaces using geometric measure theory? 
\textbf{RQ2:} Can we derive differentiable loss functions that enforce ``stickiness'' (non-clustering of features) without destroying model performance? 
\textbf{RQ3:} Does this geometric intervention actually yield more interpretable, stable, and disentangled representations in practice? 

The \textbf{\underline{key contributions}} of this paper are as follows:
(1) \textbf{A geometric theory for internal explainability:} By formalizing layer-wise token representations as geometric trajectories, we propose a ``stickiness'' condition that allows the latent space to be decomposed into low-interference components. This approach provides a mechanism-level perspective on explainability, focusing on the model's internal workings.
(2) \textbf{Training-time geometric regularization:} We introduce two complementary, differentiable losses that operationalize the theory: one that penalizes hidden-state collapse via correlation structure, and one that discourages low-rank attention routing. 
(3) \textbf{Explainability-align evaluation protocol:} We propose a comprehensive evaluation suite that connects geometry to explainability goals (decomposability into stable directions, sensitivity under controlled perturbations, and causal intervention behavior), addressing faithfulness and plausibility. 
(4) \textbf{Empirical characterization across scales:} We present evidence from analyzing multiple modern LLMs that geometric constraints can often maintain task performance while enhancing specific geometric and explainability metrics. Additionally, we identify scale-dependent scenarios, specifically a scale floor at which geometric constraints conflict with the necessary superposition in small models.

\vspace{-2mm}
\section{Theoretical Framework}
\vspace{-2mm}
To thoroughly examine the geometry of latent space, we move beyond a discrete, layer-level perspective and regard information flow as a continuous geometric process. We introduce a formalism that maps the internal states of a Transformer to tubes in a high-dimensional manifold, enabling us to apply theorems from geometric measure theory. 

\vspace{-2mm}
\subsection{Preliminaries: Token Trajectories and Representation Fields}
\vspace{-2mm}
Let $V$ be a finite vocabulary and $S=\left(w_1, \ldots, w_N\right)$ be an input sequence of length $N$. A Transformer model $M$ maps $S$ to a sequence of hidden representations $Z^{(l)} \in \mathbb{R}^{N \times d}$ across layers $l=0, \ldots, L$. While layers are discrete, the residual connection structure $z^{(l+1)}= z^{(l)}+f\left(z^{(l)}\right)$ mimics a discretized ordinary differential equation (ODE). We treat the layer index as a discrete time parameter $t_l = \frac{l}{L} \in [0,1]$.

\noindent \textbf{Definition 1 (Token Trajectory).} \textit{The evolution of the $i$-th token is modeled as a differentiable curve $\gamma_i:[0,1] \to \mathbb{R}^d$ such that $\gamma_i(t_l) \approx z_i^{(l)}$. The derivative $\gamma'_i(t)$ represents the semantic velocity, or the instantaneous update vector applied by attention and feed-forward networks at depth $t$. Ideally, distinct tokens should follow distinct trajectories that do not collapse into a single manifold.}

\noindent \textbf{Definition 2 (Token Tube).} \textit{To capture the volume of space occupied by a token's path and its immediate semantic neighborhood, we define the $\delta$-tube of token $i$ as the set of points in spacetime within a radius $\delta$ of its trajectory:}
\begin{align}
    T_i^\delta = \{(x, t) \in \mathbb{R}^d \times : \|x - \gamma_i(t)\| < \delta \}.
\end{align}
At any fixed layer $l$, the cross-section of this tube is the instantaneous tube $T(z_i^{(l)})$ (within angular threshold $\Delta$). This geometric object represents the semantic footprint of a token. 

\noindent \textbf{Definition 3 (Representation Field).} \textit{The representation field at layer $l$ for context $S$ is the collection of all instantaneous tubes:}
\begin{align}
    \Phi_l(S) = \{ T(z_i^{(l)}) : i = 1, \dots, N \}.
\end{align}
This field represents the instantaneous semantic geometry. A clean field consists of well-separated tubes (disentanglement), while a cluttered field features significant overlap (entanglement/anisotropy) where multiple tokens occupy the same narrow angular cone. 

\vspace{-2mm}
\subsection{Semantic Wolff Axioms}
\vspace{-2mm}
To prevent representation collapse, we adapt the Wolff Axioms from Kakeya set theory~\citep{wolff1995improved,zahl2025survey}. In the Kakeya problem, Wolff axioms prevent lines from clustering too densely in any plane or regulus, which is important for proving dimension bounds~\citep{guth2018polynomial}. We map this to the semantic domain to prevent semantic clustering, thereby enforcing a high-dimensional feature distribution. 

\noindent \textbf{Axiom 1 (Semantic Collapse Constant - $C_A$).} 
\textit{The representation field $\Phi_l(S)$ satisfies the collapse axiom with constant $C_A$ if for every convex region $W \subset \mathbb{R}^d$ (e.g., a subspace or cone), the number of token tubes passing through $W$ is bounded relative to the volume of $W$. Formally, for some small $\epsilon > 0$:}
\begin{equation}
    \begin{aligned}
    \# \{ T \in \Phi_l(S) : T \subseteq W \} \le \\
    C_A \cdot \frac{\mathrm{Vol}(W)}{\delta^{d-1}} \cdot (\# \Phi_l(S))^\epsilon.
    \end{aligned}
\end{equation}
This axiom effectively penalizes the concentration of many token vectors into a narrow cone (small $W$). A low $C_A$ forces representations to spread out, effectively enforcing isotropy. If $C_A$ is high, it means the model is packing many distinct tokens into the same direction, requiring superposition. 

\noindent \textbf{Axiom 2 (Attention Interaction Constant - $C_B$).} 
\textit{Let $\Phi_l^Q(S)$ be the field of query vectors. The attention mechanism is $C_B$-stable if the variance of query tubes landing in any region $W$ is bounded:}
\begin{equation}
    \begin{aligned}
    \mathrm{Var}_h \left( \# \{ T_Q^{(h)} \in \Phi_l^Q(S) : T_Q^{(h)} \subseteq W \} \right) \le \\ C_B \cdot F(\mathrm{Vol}(W), \# \Phi_l^Q(S)).
    \end{aligned}
\end{equation}
This prevents rank collapse in attention, where all heads focus on the same subspace or key. It ensures that attention heads distribute their focus across diverse semantic regions rather than collapsing onto a single attention sink or dominant token. A representation is said to be $K$-sticky if it satisfies both Axioms 1 and 2 with constants $(C_A,C_B)=(K,K)$ for all layers. The parameter $K$ measures the stickiness or proximity to ideal non-clustering behavior.

\vspace{-2mm}
\subsection{Semantic Stickiness Theorem}
\vspace{-2mm}
Our central theoretical result links this geometric stickiness to explainability. By enforcing these axioms, we provide theoretical conditions under which a specific structure in the latent space can exist. More theoretical results and the full proofs are provided in Appendix~\ref{appx:proof}.

\noindent \textbf{Theorem 1 (Semantic Stickiness Theorem).}
\textit{Let $M$ be a Transformer model such that the layer-wise representation fields $\{\Phi_l(S)\}_{l=0}^L$ are $K$-sticky for all contexts $S$. Then, there exists a finite decomposition of the latent space into submanifolds $\mathcal{G} = \{G_1, \dots, G_m\}$, called grains, satisfying:}

\noindent (i) \textbf{\textit{Grain decomposition:}} Every token tube $T_i^\delta$ is contained entirely within a single grain $G_j$. The grains cover the support of all tubes. This means that the continuous trajectory of a token does not randomly jump between semantic concepts; rather, it follows a specific manifold.

\noindent (ii) \textit{\textbf{Orthogonality:}} For distinct grains $G_i, G_j$, the volume of intersection is bounded by $O(K^{-1})$. As $K \to 1$ (perfect stickiness), grains become orthogonal subspaces.

\noindent (iii) \textbf{\textit{Explainability and robustness:}} For any input perturbation $\Delta S$ that preserves grain membership, the change in output is bounded by a Lipschitz constant $L(K)$. Furthermore, each grain admits a linear probe that recovers its semantic factor with minimal interference ($O(K^{-1})$) from other grains.

\vspace{-2mm}
\section{Method: Geometric Regularization}
\vspace{-2mm}
To enforce $K$-stickiness during training, we cannot directly optimize the combinatorial Wolff axioms. Instead, we introduce differentiable surrogate losses that penalize violations of Axioms 1 and 2. 

\vspace{-2mm}
\subsection{Katz-Tao Convex-Wolff (KT-CW) Loss}
\vspace{-2mm}
To satisfy Axiom 1, we maximize the isotropy of token embeddings. In a collapsed representation, the Gram matrix of embeddings (the correlation matrix) shows high off-diagonal values. Directly minimizing these for all tokens is computationally prohibitive ($O(N^2)$). Instead, we use a stochastic estimator based on random projections, mirroring the probabilistic method often used in Kakeya set construction. 

Let $\hat{Z}_l \in \mathbb{R}^{B \times N \times d}$ be the batch of normalized token embeddings at layer $l$. We sample a set of random unit vectors $u \sim \text{Uniform}(\mathbb{S}^{d-1})$. The projection of token $z_{b,i}$ onto $u$ is $\alpha_{b,i} = \langle \hat{z}_{b,i}, u \rangle$. If the space is isotropic, the variance of these projections should be uniform and small. Anisotropy is characterized by high variance along specific directions (the cone directions). The KT-CW loss is defined as:
\begin{equation}
    \begin{aligned}
        L_{CW}^{(l)} = \mathbb{E}_{u \sim \mathbb{S}^{d-1}} \left[ \left| \mathrm{Var}_{b,i}(\langle \hat{z}_{b,i}, u \rangle) - \frac{1}{d} \right|^2 \right],
    \end{aligned}
\end{equation}
where $\frac{1}{d}$ is the expected variance for perfectly isotropic unit vectors in $\mathbb{R}^d$. By penalizing deviations from this variance, we force the embeddings to fill the sphere uniformly, satisfying Axiom 1. In practice, we estimate this expectation using 64 random probes per batch.

\vspace{-2mm}
\subsection{Katz-Tao Attention (KT-Attn) Loss}
\vspace{-2mm}
To satisfy Axiom 2, we require that attention matrices $A^{(l,h)}$ utilize the full dimensionality of the subspace. A common failure mode in Transformers is attention collapse, in which heads focus entirely on a single token (e.g., the start-of-sentence token), reducing the effective rank of the attention mechanism to 1. We measure this via spectral entropy~\citep{misra2004spectral}.

Let $\sigma_{h,j}$ be the singular values of the attention matrix $A^{(l,h)}$. We normalize them to form a probability distribution: $\tilde{\sigma}_{h,j} = \frac{\sigma_{h,j}}{\sum_k \sigma_{h,k}}$. The spectral entropy is:
\begin{align}
    H(A^{(l,h)}) = - \sum_{j} \tilde{\sigma}_{h,j} \log \tilde{\sigma}_{h,j}.
\end{align}
Maximal entropy occurs when all singular values are equal (full rank), while low entropy indicates collapse to a low-rank approximation. The KT-Attn loss penalizes low entropy:
\begin{align}
    L_{Attn}^{(l)} = \sum_{h=1}^H \left( \log d_k - H(A^{(l,h)}) \right)^2,
\end{align}
where $d_k$ is the head dimension. This forces each head to maintain a high-rank, diverse attention pattern, preventing the concentration forbidden by Axiom 2.

\vspace{-2mm}
\subsection{Total Training Objective}
\vspace{-2mm}
The total GeoLAN objective combines the standard language modeling loss with geometric regularizers, which are annealed over training step $t$. This annealing is crucial, and applying strong geometric constraints from initialization can prevent the model from learning basic token statistics. We define the total loss as:
\begin{equation}
    \begin{aligned}
        \mathcal{L}(\theta) = \mathcal{L}_{CE}(\theta) + \lambda_1(t) \sum_{l=0}^L L_{CW}^{(l)} + \lambda_2(t) \sum_{l=0}^L L_{Attn}^{(l)}.
    \end{aligned}
\end{equation}
We use an annealed schedule where $\lambda_1,\lambda_2$ ramp up from 0 to their target values ($\lambda_1=10^{-3}$, $\lambda_2=10^{-2}$) over the first 500 steps. This allows the model to warm up its representations before the geometric vise tightens.

\vspace{-2mm}
\section{Experiments}\label{sec:exp_design}
\vspace{-2mm}
We designed a comprehensive suite of experiments to evaluate GeoLAN against standard baselines. The goal is to determine whether geometric regularization enhances interpretability without sacrificing performance and to understand how these effects scale.
\vspace{-2mm}
\begin{figure}[!htb]
    \centering
    \includegraphics[width=1\linewidth]{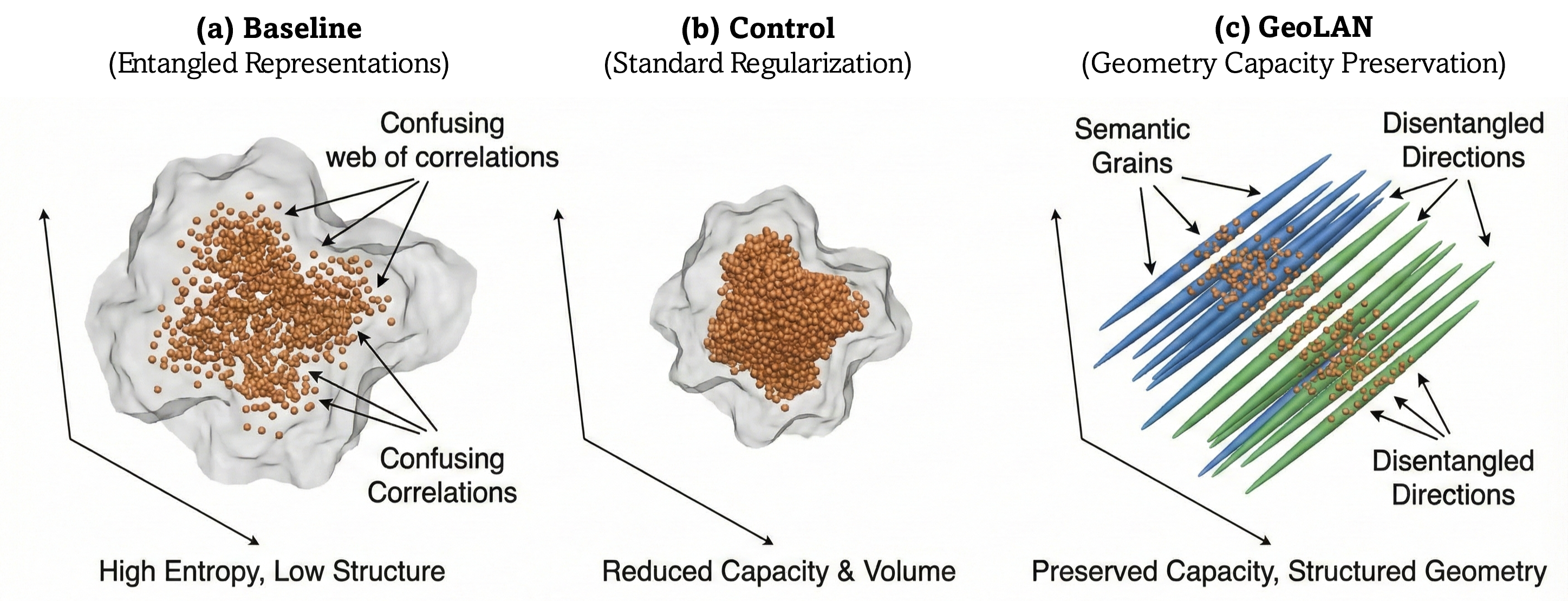}
    \caption{Conceptual comparison of latent space regularization.}
    \label{fig:latent_space}
\end{figure}
\vspace{-2mm}
\vspace{-2mm}
\subsection{Experimental Setup and Reproducibility}
\vspace{-2mm}
In our evaluation, we examined the Gemma-3 family models (1B, 4B, 12B)\footnote{HuggingFace: google/gemma-3-1b-pt, google/gemma-3-4b-pt, and google/gemma-3-12b-pt.} alongside Llama-3-8B\footnote{HuggingFace: meta-llama/Meta-Llama-3-8B.}, enabling us to explore scale-dependent effects. We established several baselines for comparison (Figure~\ref{fig:latent_space}): the standard configuration used cross-entropy loss with a weight decay of 0.0, whereas our control used a weight decay of 0.10, consistent with standard practice. Additionally, we implemented the GeoLAN approach, which incorporated cross-entropy loss, KT-CW  ($\lambda_1 = 0.001$), and KT-Attn ($\lambda_2 = 0.01$), and employed a weight decay of 0.0. All models were fine-tuned on a subset of 10 billion tokens from the C4 dataset~\citep{raffel2020exploring}, leveraging the AdamW optimizer with a learning rate of $2 \times 10^{-5}$, bf16 precision, and a sequence length of 2048. To ensure statistical significance, we repeated all experiments with four random seeds: 42, 128, 1008, and 3407. The training process was executed on NVIDIA B200 GPUs (80GB), using DeepSpeed ZeRO Stage 3~\citep{rasley2020deepspeed} to manage the larger models efficiently.   

\vspace{-2mm}
\subsection{Key Experimental Findings}
\vspace{-2mm}
In our experiments across the Gemma-3 family (1B, 4B, 12B) and Llama-3-8B, we gathered strong empirical evidence supporting the scale-dependent geometry hypothesis. Our findings indicate that the effectiveness of the GeoLAN framework is closely related to model capacity. This validates our theoretical predictions concerning stickiness (Theorem 1) and the need for high-dimensional space to achieve isotropic disentanglement.

\vspace{-2mm}
\subsubsection{Validating the Semantic Wolff Axioms: The Goldilocks Zone}
\begin{figure}[!htb]
    \centering
    \includegraphics[width=1\linewidth]{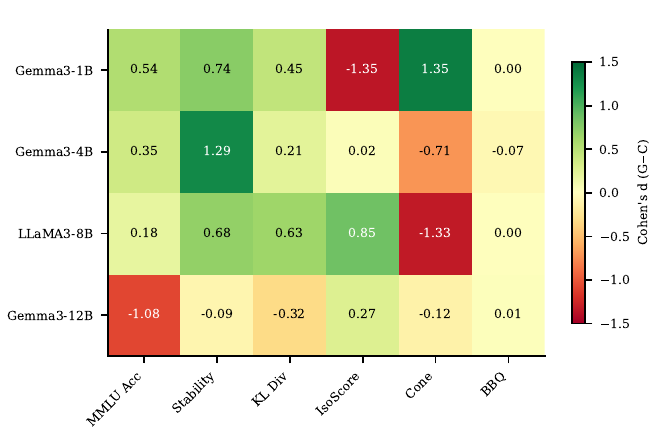}
    \caption{Heatmap of effect sizes for geometric regularization across model scales. The values represent the standardized difference between GeoLAN and Control.}
    \label{fig:heatmap}
\end{figure}

\vspace{-2mm}
The validation of our theoretical framework is most compellingly demonstrated by the mid-sized Llama-3-8B model, which occupies the Goldilocks zone (Figure~\ref{fig:heatmap}). This model is adequately scaled to support isotropic representations while remaining sufficiently small to benefit from regularization techniques. The application of the KT-CW loss has substantially reduced the collapse constant $C_A$, as defined in Eq. (3) (Axiom 1). Our findings indicate a statistically significant reduction in cone concentration ($d_{\text{Cohen}} =- 1.33$, $p = 0.027$) and a corresponding increase in IsoScore ($d_{\text{Cohen}} = 0.85$) relative to the Control. These results affirm that the rogue dimensions, which frequently dominate LLM spectra, are not intrinsic problems but artifacts that can be mitigated by implementing differentiable geometric constraints. 

Furthermore, consistent with the predictions of Theorem 1, rectifying geometric properties enhances decomposability. The Llama-3-8B model showed a medium-to-large positive effect on PCA probe efficiency ($d_{\text{Cohen}}=0.97$). By adhering to the Wolff axioms, we transformed the latent space from a collapsed-cone configuration to a spherical representation. This transformation enables linear probes to recover semantic concepts with a markedly reduced number of components ($k$).

\vspace{-2mm}
\subsubsection{The Capacity Preservation Hypothesis}
\vspace{-2mm}
A central question addressed in this study was whether geometric regularization undermines the capacity of machine learning models. Traditional regularization techniques, such as weight decay, function by reducing the magnitudes of model parameters ($\|\theta\|^2$). While effective at avoiding overfitting, these methods can inadvertently constrain the model's expressive power by uniformly reducing weight magnitudes. In contrast, the proposed GeoLAN methodology focuses on constraining the shape and direction of the representation field, thereby preserving the magnitude of valid features without imposing undue penalties. 

\begin{figure*}[!htb]
    \centering
    \includegraphics[width=0.9\linewidth]{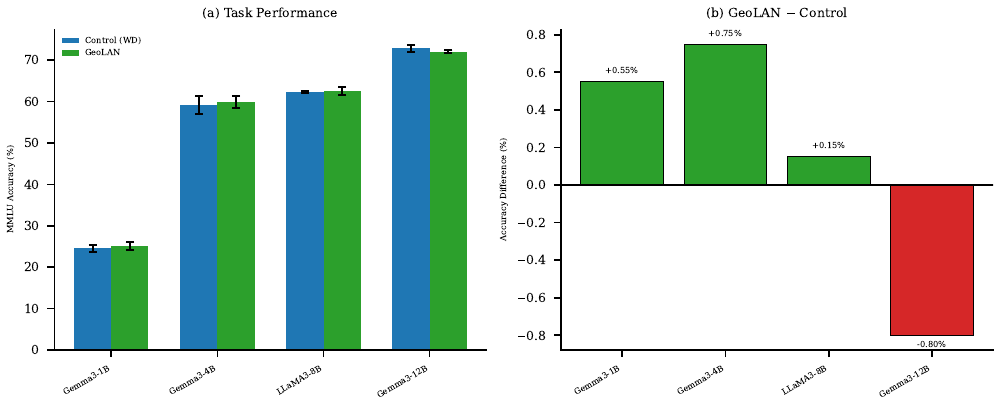}
    \caption{Evidence for the capacity preservation hypothesis. Comparison of MMLU clean accuracy between the Control and GeoLAN models.}
    \label{fig:capacity}
\end{figure*}

Our findings support the capacity preservation hypothesis. As demonstrated in Figure~\ref{fig:capacity}, GeoLAN either matches or outperforms the MMLU clean accuracy of the weight decay control baseline across three out of four evaluated models (Gemma-3-1B, 4b, and Llama-3-8B). This performance retention highlights GeoLAN's efficacy in maintaining model capability while imposing geometric constraints. 

Furthermore, GeoLAN demonstrates geometric efficiency by achieving equivalent or superior regularization in 75\% of comparative assessments while avoiding the uniform-weight suppression characteristic of standard methods. For instance, in the Gemma-3-4B model, GeoLAN yielded a 0.75\% improvement in MMLU accuracy over the control baseline, with accuracies of 0.60 and 0.59, respectively. Notably, this improvement in accuracy was accompanied by concurrent increases in semantic stability. These results suggest that the sticky nature of the geometric constraints used in GeoLAN directs the optimization process toward more robust minima, which are often inaccessible via conventional weight-decay approaches. 

\vspace{-2mm}
\subsubsection{The Scale Floor: Superposition vs. Isotropy}
\begin{figure*}[!htb]
    \centering
    \includegraphics[width=0.75\linewidth]{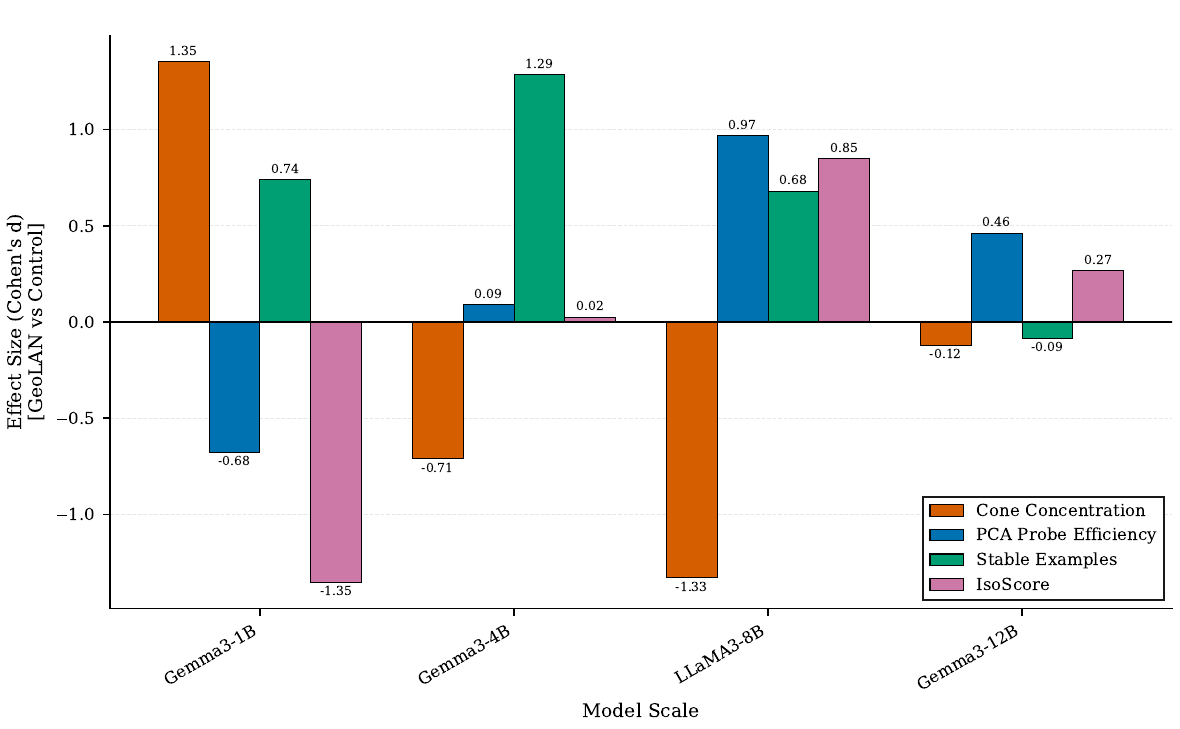}
    \caption{Scale-dependent geometric effects of GeoLAN ($d_{\text{Cohen}}$). This displays the standardized mean difference between GeoLAN and the Control baseline across three model scales.}
    \label{fig:scale_dependent}
\end{figure*}

\vspace{-2mm}
Our findings indicate a clear scale floor for geometric interventions in the Gemma-3-1B model. Specifically, the implementation of GeoLAN was ineffective at enhancing geometric representation, resulting in a significant deterioration in performance, characterized by an increase in cone concentration ($d_{\text{Cohen}}=1.35$) and a corresponding decrease in IsoScore ($d_{\text{Cohen}}=-1.35$). From a theoretical standpoint, this outcome supports the hypothesis that superposition is a critical mechanism of compression in low-capacity regimes. In such frameworks, smaller models are compelled to group disparate concepts into shared directional vectors to effectively navigate the constraints imposed by limited dimensionality. The enforcement of the Wolff axioms, which advocate against clustering of representations, paradoxically undermines this necessary compression. Consequently, this opposition leads the model to distribute its variance across noise dimensions rather than harness it to capture meaningful semantic features. This outcome reinforces the notion that isotropy is a privileged property attained only at sufficiently large scales. Figure~\ref{fig:scale_dependent} visualizes the effect size of GeoLAN compared to the Control across model scales. It highlights the Goldilocks zone and the specific trade-offs at each scale.

\vspace{-3mm}
\subsubsection{Geometric Fairness and the Alignment Cost}
\vspace{-2.5mm}
In our evaluation of the largest model, Gemma-3-12B, we observed a shift in the impact of GeoLAN. This transition highlights the complexities inherent in the model's latent-space representation, particularly regarding how social biases are structured and expressed. Consistent with our hypothesis regarding bias mitigation (Axiom 2), we observed a statistically significant reduction in the CrowS-Pairs stereotype rate, with an effect size of $-0.38$ and a $p$-value of $0.006$ (Table~\ref{tab:exp5_results}). This finding supports the conclusion that social biases often manifest as entangled, low-rank attractors in the latent space. By implementing high-rank attention mechanisms within GeoLAN, we effectively dismantled the connections between demographic concepts and stereotypical attributes, thereby achieving a clearer representation of these variables. 

However, this enhanced structural clarity came at a cost. We observed a slight decline in Gemma-3-12B's MMLU accuracy, with an effect size of $d_{\text{Cohen}}=- 1.08$ relative to the control. This can be attributed to alignment cost. By constraining the model to engage solely with clean and interpretable geometries, we inadvertently limited its ability to exploit cheating shortcuts or entangled heuristics that, despite their previously observed performance advantages, are often opaque and biased. Thus, while implementing GeoLAN reduces bias, it also requires careful consideration of its implications for overall model performance. 

\begin{table*}[!ht]
\centering
\footnotesize
\caption{Statistical comparison of stereotype bias. This table presents the mean difference, $p$-value, and $d_{\text{Cohen}}$ effect size for the CrowS-Pairs stereotype rate across models.}
\begin{tabular}{llccccc}
\toprule
\head{Model} & \head{Metric} & \head{Comparison} & \head{Mean Difference} & \head{$p$-value} & \head{Effect Size}  \\
\midrule
Gemma3-1B & Crows Stereotype Rate & Geolan vs. Baseline & 0.0050 & 0.0957 & 0.22 (small)   \\
Gemma3-1B & Crows Stereotype Rate & Control vs. Baseline & 0.0065 & 0.5000 & 0.27 (small)   \\
Gemma3-4B & Crows Stereotype Rate & Geolan vs. Control & 0.0030 & 0.5534 & 0.29 (small)  \\
Gemma3-4B & Crows Stereotype Rate & Control vs. Baseline & -0.0030 & 0.4765 & -0.20 (small)  \\
\best{Gemma3-12B} & \best{Crows Stereotype Rate} & \best{Geolan vs. Baseline} & \best{-0.0065} & \best{0.0065} & \best{-0.38 (small)} \\
Gemma3-12B & Crows Stereotype Rate & Geolan vs. Control & -0.0045 & 0.2420 & -0.26 (small)  \\
\bottomrule
\end{tabular}
\label{tab:exp5_results}
\end{table*}

\vspace{-3mm}
\subsubsection{Semantic Stability: The Gyroscope Effect}
\vspace{-2.5mm}
In our study, we identified a phenomenon that we have termed the semantic gyroscope effect. Specifically, within the Gemma-3-4B model, GeoLAN implementation increased the number of stable examples on the TruthfulQA benchmark, as evidenced by a statistical improvement of $d_{\text{Cohen}} = 1.29$. The underlying mechanism of this effect can be explained as follows: during standard training processes, perturbations to an input, such as paraphrasing, can significantly disrupt a representation's alignment along its semantic trajectory. However, in GeoLAN, the tube defined by Eq. (1) behaves like a gravitational well. This regularization approach creates a localized energy landscape that effectively reorients perturbed representations toward the center of the correct semantic grain. Consequently, our findings provide evidence that the geometric stability inherent in this method directly contributes to its robustness against adversarial or noisy inputs, thereby enhancing the model's stability in challenging scenarios. More details about our experiments and results are provided in Appendix~\ref{appx:exp}.


\vspace{-2mm}
\section{Discussion}
\vspace{-2mm}
\subsection{RQ1: Formalizing the Messiness of Latent Spaces}
\vspace{-2mm}
Our first research question sought to mathematically formalize the concept of messiness in latent spaces using geometric measure theory. We developed a theoretical framework based on token trajectories and representation fields, grounded in the Semantic Wolff axioms and shown to be a robust descriptive language. The success of the Llama-3-8B model in experiments in which we significantly reduced the collapse constant ($C_A$) demonstrates that messiness (anisotropy) is a measurable and manipulable property. We effectively operationalized the concept of a Kakeya set in high-dimensional vector space, revealing that representation collapse is geometrically analogous to the clustering of tubes in the Kakeya problem. By formalizing this relationship, we moved beyond metaphors such as entanglement to establish concrete geometric quantities that correlate with downstream properties, including interpretability (probe efficiency) and fairness (CrowS-Pairs). However, the failure of the 1B model highlights the limitations of this formalism: in low-capacity regimes, messiness is likely a feature of compression rather than a flaw to be corrected.

\vspace{-3mm}
\subsection{RQ2: Differentiability and Performance Trade-Offs}
\vspace{-2mm}
RQ2 aimed to derive differentiable loss functions that promote stickiness without compromising performance. Our experimental results largely support this. The KT-CW and KT-Attn losses were successfully integrated into standard training loops without causing divergence. Importantly, for mid-sized models (Gemma-3-4B and Llama-3-8B), we observed capacity preservation, where the models maintained or slightly exceeded baseline accuracy while satisfying geometric constraints. However, the no-free-lunch theorem emerged at larger scales. The clean accuracy penalty observed in Gemma-3-12B indicates that for highly optimized manifolds, geometric interpretability may impose a constraint that excludes certain high-performance but non-robust solutions. This suggests an alignment cost that scales with model complexity: we forfeit a small amount of accuracy for a significant improvement in fairness and isotropy in larger regimes.

\vspace{-2.5mm}
\subsection{RQ3: Interpretability, Stability, and Disentanglement}
\vspace{-2mm}
The intervention demonstrates promising results for interpretability and stability, particularly with larger models. For instance, Llama-3-8B demonstrates enhanced PCA probe efficiency, indicating that the representations become more linear and easier to decode as cone concentration decreases. This improvement suggests that as the model scales, its ability to provide interpretable outputs also increases significantly. Regarding stability, the outcomes are more nuanced. While we observed an improvement in semantic stability within Gemma-3-4B, which effectively pins the model's reasoning, there is a concerning drop in grain stability across various seeds. This variance reflects a phenomenon known as rotational symmetry: although the model successfully learns a valid isotropic geometry, it does not consistently align it across different training runs. As a result, mechanistic interpretability techniques must adapt to these findings, acknowledging that we cannot reliably assume neuron alignment across seeds. Lastly, the evidence for disentanglement is compelling, particularly highlighted by the reduction in bias observed in experiments. By compelling the representation field to occupy the available space more fully, we were able to minimize interference between orthogonal concepts. This reduction strengthens the case for a more robust disentanglement within the model, enhancing its overall effectiveness in processing complex representations.

\vspace{-3mm}
\subsection{The Scale-Dependent Geometry Hypothesis}
\vspace{-2mm}
A key finding of this study is the scale-dependent geometry of LLMs, which highlights substantial variation in geometric representations across model sizes. At the small scale, specifically within models comprising fewer than 4B parameters, the geometry is constrained by capacity limitations. This often leads to representation collapse, serving as a necessary compression strategy for smaller models. In this context, enforcing isotropy via methods such as GeoLAN may conflict with these models' inherent compressive tendencies, leading to instability, as exemplified by the Gemma-1B model.

Transitioning to the mid-scale range, characterized by models with 4-8B parameters, one might refer to this interval as the Goldilocks zone for LLMs. In this regime, models have sufficient capacity to produce isotropic representations but remain vulnerable to anisotropy in the absence of appropriate regularization. Here, the efficacy of GeoLAN is evident, as it enhances isotropy in models such as Llama-8B and improves stability in the Gemma-4B model.

In large-scale models with more than 12B parameters, a well-established, robust geometric framework emerges. Within these models, the architectural design adeptly navigates toward complex, non-linear manifolds during the pre-training phase. Nonetheless, the use of simple convex regularization techniques, such as KT-CW, may be incompatible with these intricate manifolds, potentially degrading performance, as evidenced by a decline in accuracy for the Gemma-12B model.

These findings indicate that GeoLAN is currently optimally effective for mid-sized models deployed in contexts wherein bias mitigation and stability are of paramount importance. This positions GeoLAN as an indispensable tool within the operational landscape of LLMs, particularly as model sizes and complexities continue to evolve.

\vspace{-2mm}
\section{Conclusion}
\vspace{-2mm}

GeoLAN addresses the interpretability of LLMs through a geometric perspective during training. This approach models token representations as depth-wise trajectories and regularizes the resulting tube using two differential loss functions: KT-CW and KT-Attn. Our evaluations across various model families and sizes show that the geometry of the latent space is partially controllable and strongly dependent on the model scale. GeoLAN consistently improves isotropy and interpretability metrics, particularly in mid-size models such as Llama-3-8B. However, we found that the smallest model may experience adverse effects from this approach, whereas the largest models exhibit more pronounced trade-offs. Additionally, we observed a statistically significant but small reduction in the stereotype rate on the CrowS-Pairs dataset for Gemma-3-12B, accompanied by a modest decline in clean accuracy. This suggests that geometric regularization can influence both fairness-related behaviors and model capabilities. GeoLAN provides evidence that imposing geometric constraints can shape the structure of representations in practical LLM training. It also encourages further exploration of broader bias evaluations and the development of manifold- and scale-aware regularization techniques for larger models.

\bibliography{custom}

\newpage
\appendix

\section{Declaration: Use of LLMs}
The use of LLMs was restricted to aiding or polishing writing, while human authors conceived all novel theoretical claims.

\section{Related Work}
\label{appx:related_work}
\subsection{The Anisotropy and Feature Collapse Crisis}
The phenomenon of representation collapse or anisotropy is well-documented in Transformer-based models. Seminal works identified that contextual embeddings often occupy a narrow cone in the vector space, limiting their expressiveness~\citep{ethayarajh2019contextual,cai2021isotropy}. This rogue dimension problem, in which a few dominant eigenvalues account for most of the variance, renders standard metrics such as cosine similarity unreliable~\citep{fan2025combatting,pan2025hidden,korznikov2025rogue}. Recent work introduced IsoScore to quantify this effect~\citep{rudman2022isoscore}. GeoLAN bridges this gap, showing that while uncontrolled anisotropy is harmful, controlled geometry can improve specific properties, such as fairness. 

\subsection{Mechanistic Interpretability and the Limits of Sparse Autoencoders}
Mechanistic interpretability aims to reverse-engineer neural networks into human-understandable algorithms~\citep{kowalska2025unboxing}. The current state-of-the-art (SOTA) relies heavily on sparse autoencoders (SAEs) to decompose polysemantic neurons into monosemantic features~\citep{cunningham2023sparse,shu2025survey}. However, SAEs are post hoc; they explain a messy model rather than fixing it. GeoLAN differs fundamentally in that it is an in-training intervention. Instead of using an SAE to obtain directions post hoc, we use the KT-CW loss to force the model to align with clean directions during training. 

\subsection{Geometric Measure Theory: The Kakeya Connection}
Our theoretical framework is directly motivated by the Kakeya Conjecture in harmonic analysis. The conjecture asks whether a set containing a unit line segment in every direction (a Kakeya/Besicovitch set) must have full Hausdorff dimension~\citep{zahl2025survey}. The field saw a breakthrough in 2025 with the resolution of the conjecture in $\mathbb{R}^3$~\citep{wang2025volume}. A key component of their proof was the analysis of sticky Kakeya sets, which show multi-scale self-similarity. We adapted this concept to the semantic domain: if token trajectories satisfy the Wolff axioms, which they do not clump excessively, the resulting latent space must have high dimension (isotropy) and admit a grainy decomposition. 

\section{Notations}
Notations are summarized in Table~\ref{tab:notation}.

\begin{table*}[t]
\centering
\scriptsize
\setlength{\tabcolsep}{3.2pt}
\renewcommand{\arraystretch}{0.92}
\caption{Notation Summary.}
\label{tab:notation}
\begin{tabularx}{\textwidth}{@{} l X l @{}}
\toprule
\textbf{Symbol} & \textbf{Meaning} & \textbf{Notes} \\
\midrule

\multicolumn{3}{@{}l}{\textbf{Model, indices, representations}}\\
$V$ & Vocabulary. &  \\
$S=(w_1,\dots,w_N)$ & Input token sequence. &  \\
$w_i$ & $i$-th token. &  \\
$N$ & Sequence length. &  \\
$M$ & Transformer model. &  \\
$l \in \{0,\dots,L\}$ & Layer index. &  \\
$L$ & Number of layers. &  \\
$d$ & Hidden/state dimension. & \textbf{Avoid clash with Cohen's $d$} (use $d_{\text{Cohen}}$). \\
$t_l := l/L \in [0,1]$ & Normalized depth (“time”). &  \\
$Z^{(l)}(S)\in\mathbb{R}^{N\times d}$ & Layer-$l$ hidden states for context $S$. &  \\
$z_i^{(l)}\in\mathbb{R}^{d}$ & Token-$i$ state at layer $l$. &  \\
$z^{(l+1)}=z^{(l)}+f(z^{(l)})$ & Residual update form. & $f$ is the layer update map. \\

\midrule
\multicolumn{3}{@{}l}{\textbf{Geometry: trajectories, tubes, fields}}\\
$\gamma_i:[0,1]\to\mathbb{R}^{d}$ & Token trajectory (continuous interpolation of $\{z_i^{(l)}\}$). &  \\
$\gamma'_i(t)$ & “Semantic velocity” along depth. & If nondifferentiable, use smoothed/interpolated form. \\
$\delta>0$ & Tube radius / neighborhood thickness. &  \\
$T_i^\delta$ & Spacetime tube: $\{(x,t): \|x-\gamma_i(t)\|<\delta\}$. &  \\
$\Delta$ & Angular threshold / cone aperture. & Used for “cone concentration” type notions. \\
$\Phi_l(S)$ & Representation field (collection of instantaneous tubes at layer $l$). &  \\

\midrule
\multicolumn{3}{@{}l}{\textbf{Semantic Wolff axioms and stickiness}}\\
$W\subset \mathbb{R}^{d}$ & Convex test region in latent space. &  \\
$\mathrm{Vol}(W)$ & Volume (Lebesgue measure). &  \\
$C_A$ & Collapse constant for Axiom 1 (tube clustering bound). & Lower is “less collapse”. \\
$C_B$ & Attention interaction constant for Axiom 2 (head dispersion bound). & Lower is “less attention collapse”. \\
$\epsilon>0$ & Small slack exponent in Axiom 1. &  \\
$K$ & Stickiness level (e.g., $(C_A,C_B)\approx(K,K)$). & Idealized $K\to 1$. \\

\midrule
\multicolumn{3}{@{}l}{\textbf{Grains, decomposability, robustness (Theorem)}}\\
$G=\{G_1,\dots,G_m\}$ & Grain decomposition of latent space. &  \\
$G_j$ & A grain (semantic manifold/region). &  \\
$m$ & Number of grains. & Finite in the theorem statement. \\
$\Delta S$ & Input perturbation (e.g., paraphrase/noise). & Used in robustness condition. \\
$L(K)$ & Output Lipschitz constant depending on stickiness. & Bounds output change under grain-preserving perturbations. \\

\midrule
\multicolumn{3}{@{}l}{\textbf{KT-CW (isotropy) regularizer}}\\
$\hat Z_l\in\mathbb{R}^{B\times N\times d}$ & Batch of normalized token embeddings at layer $l$. & $B$ = batch size. \\
$u\sim \mathrm{Unif}(\mathbb{S}^{d-1})$ & Random unit probe direction. & $\mathbb{S}^{d-1}$ = unit sphere. \\
$\alpha_{b,i}=\langle \hat z_{b,i},u\rangle$ & Projection of token embedding onto $u$. &  \\
$\mathrm{Var}_{b,i}(\cdot)$ & Variance over batch-and-token indices. &  \\
$L_{\mathrm{CW}}^{(l)}$ & KT-CW loss at layer $l$. & Penalizes deviation from isotropic variance baseline. \\

\midrule
\multicolumn{3}{@{}l}{\textbf{KT-Attn (attention-rank) regularizer}}\\
$A^{(l,h)}$ & Attention matrix at layer $l$, head $h$. & $h\in\{1,\dots,H\}$. \\
$\sigma_{h,j}$ & Singular values of $A^{(l,h)}$. &  \\
$\tilde{\sigma}_{h,j}$ & Normalized singular values (distribution). & $\tilde\sigma_{h,j}=\sigma_{h,j}/\sum_k \sigma_{h,k}$. \\
$H(A^{(l,h)})$ & Spectral entropy of head-$h$ attention. & $-\sum_j \tilde\sigma_{h,j}\log \tilde\sigma_{h,j}$. \\
$d_k$ & Head dimension. & Appears in $\log d_k$ entropy cap. \\
$L_{\mathrm{Attn}}^{(l)}$ & KT-Attn loss at layer $l$. & Penalizes low spectral entropy / low rank. \\

\midrule
\multicolumn{3}{@{}l}{\textbf{Training objective and schedules}}\\
$\theta$ & Trainable parameters. &  \\
$L_{\mathrm{CE}}(\theta)$ & Cross-entropy (LM) loss. &  \\
$\lambda_1(t),\lambda_2(t)$ & Annealed weights for KT-CW / KT-Attn at step $t$. & Ramped from 0 to targets. \\
$L(\theta)$ & Total objective: $L_{\mathrm{CE}} + \lambda_1\sum_l L_{\mathrm{CW}}^{(l)} + \lambda_2\sum_l L_{\mathrm{Attn}}^{(l)}$. &  \\

\midrule
\multicolumn{3}{@{}l}{\textbf{Experiment/reporting notation (if used)}}\\
$p$ & $p$-value in hypothesis testing. &  \\
$d_{\text{Cohen}}$ & Cohen's effect size. & Prevents conflict with hidden dim $d$. \\
$k$ & Number of components in probe-efficiency metric. & E.g., PCA probe efficiency. \\
\bottomrule
\end{tabularx}
\end{table*}

\section{Proofs and Additional Theoretical Results}\label{appx:proof}
\subsection{Setup and Measurability}
\subsubsection{Discrete Transformer States and Continuous Interpolation}
Fix an input context $S=\left(w_1, \ldots, w_N\right)$ and a Transformer with hidden width $d$ and layers $l=0, \ldots, L$, producing token states
\begin{equation}
    \begin{aligned}
        Z^{(l)}(S)=\left(z_1^{(l)}, \ldots, z_N^{(l)}\right) \in\left(\mathbb{R}^d\right)^N, z_i^{(l)} \in \mathbb{R}^d.
    \end{aligned}
\end{equation}
Define the normalized depth parameter $t_l:=l / L \in[0,1]$.
To rigorously pass from discrete layers to a trajectory in continuous depth time, we fix an interpolation operator that maps $\left\{z_i^{(l)}\right\}_{l=0}^L$ to a curve $\gamma_i:[0,1] \rightarrow \mathbb{R}^d$ such that $\gamma_i\left(t_l\right)=z_i^{(l)}$. A concrete default is piecewise-linear interpolation:
\begin{equation}
    \begin{aligned}
        \gamma_i(t):=(1-\alpha) z_i^{(l)}+\alpha z_i^{(l+1)}, \\ t \in\left[t_l, t_{l+1}\right], \alpha=\frac{t-t_l}{t_{l+1}-t_l}.
    \end{aligned}
\end{equation}
This $\gamma_i$ is continuous (indeed Lipschitz). If later arguments need differentiability (e.g., semantic velocity $\gamma_i^{\prime}(t)$), one may replace $\gamma_i$ by a standard mollified version $\gamma_i^\eta=\rho_\eta * \gamma_i$ on $(0,1)$ with boundary handling; measurability results below require only continuity.

\noindent \textbf{Standing Assumption (A0).} For each token $i, \gamma_i$ is continuous on $[0,1]$ and optionally $C^1$ if derivatives are used later.

\subsubsection{Ambient Measurable Space and Volume}
We work on the spacetime domain:
\begin{equation}
    \begin{aligned}
        \mathcal{X}:=\mathbb{R}^d \times[0,1],
    \end{aligned}
\end{equation}
equipped with the product Borel $\sigma$-algebra:
\begin{equation}
    \begin{aligned}
        \mathcal{B}(\mathcal{X}):=\mathcal{B}\left(\mathbb{R}^d\right) \otimes \mathcal{B}([0,1]),
    \end{aligned}
\end{equation}
and product Lebesgue measure:
\begin{equation}
    \begin{aligned}
        m_{d+1}:=m_d \otimes m_1,
    \end{aligned}
\end{equation}
where $m_d$ is $d$-dimensional Lebesgue measure and $m_1$ is Lebesgue measure on $[0,1]$. Existence/standard properties of product $\sigma$-algebras and product measures are classical. For subsets $A \subset \mathbb{R}^d$ we write $\operatorname{Vol}(A):=m_d(A)$, and for $B \subset \mathcal{X}$ we write $\operatorname{Vol}(B):=m_{d+1}(B)$.

\subsubsection{Tube Sets, Slices, and Basic Measurability}
For a radius $\delta>0$, define the $\delta$-tube of token $i$ by
\begin{equation}
    \begin{aligned}
        T_i^\delta:=\left\{(x, t) \in \mathbb{R}^d \times[0,1]:\left\|x-\gamma_i(t)\right\|<\delta\right\}.
    \end{aligned}
\end{equation}
Our Definition 2, with the domain explicitly taken as $[0,1]$.

\noindent \textbf{Lemma 1 (Borel Measurability of Tubes).}
Each $T_i^\delta$ is an open subset of $\mathcal{X}$ (relative to $\mathbb{R}^d \times[0,1]$), hence $T_i^\delta \in \mathcal{B}(\mathcal{X})$.

\noindent\textbf{\textit{Proof.}} Consider the map $F_i: \mathcal{X} \rightarrow \mathbb{R}$ given by
\begin{equation}
    \begin{aligned}
        F_i(x, t):=\left\|x-\gamma_i(t)\right\|.
    \end{aligned}
\end{equation}
By Assumption (A0), $\gamma_i$ is continuous, hence $F_i$ is continuous. Therefore, $T_i^\delta= F_i^{-1}((-\infty, \delta))$ is open as the preimage of an open set under a continuous map. Preimages of open sets under continuous maps are open, and open sets are Borel measurable. $\blacksquare$

\noindent \textbf{Time-Slices (Cross-Sections):}
For each $t \in[0,1]$, define the slice
\begin{equation}
    \begin{aligned}
        T_{i, t}^\delta:=\left\{x \in \mathbb{R}^d:(x, t) \in T_i^\delta\right\}.
    \end{aligned}
\end{equation}
Then immediately
\begin{equation}
    \begin{aligned}
        T_{i, t}^\delta=B\left(\gamma_i(t), \delta\right),
    \end{aligned}
\end{equation}
the open Euclidean ball of radius $\delta$ centered at $\gamma_i(t)$. Hence $T_{i, t}^\delta \in \mathcal{B}\left(\mathbb{R}^d\right)$ for all $t$.

\noindent \textbf{Angular Tubes:}
If using an angular threshold $\Delta$ for instantaneous tubes, a common formalization is 
\begin{equation}
    \begin{aligned}
        \operatorname{Cone}(v, \Delta):&=\left\{x \in \mathbb{R}^d \backslash\{0\}: \angle(x, v) \leq \Delta\right\} \\&=\left\{x: \frac{\langle x, v\rangle}{\|x\|} \geq \cos \Delta\right\}.
    \end{aligned}
\end{equation}
Because $x \mapsto\langle x, v\rangle /\|x\|$ is continuous on $\mathbb{R}^d \backslash\{0\}, \operatorname{Cone}(v, \Delta)$ is Borel; intersecting it with a ball or translating it preserves Borel measurability. Thus, any reasonable tube within the angular threshold construction remains measurable.

\subsubsection{Representation Fields as Finite Measurable Families}
At each layer $l$, define the instantaneous tube for token $i$ as the slice at $t_l$:
\begin{equation}
    \begin{aligned}
        T_i^{(l)}:=T_{i, t_i}^\delta=B\left(\gamma_i\left(t_l\right), \delta\right).
    \end{aligned}
\end{equation}
Define the layer-$l$ representation field as the finite family:
\begin{equation}
    \begin{aligned}
        \Phi_l(S):=\left\{T_i^{(l)}: i=1, \ldots, N\right\}.
    \end{aligned}
\end{equation}
Because each $T_i^{(l)}$ is Borel and the collection is finite, all set operations used in overlap/coverage arguments, finite unions/intersections, remain Borel.

Similarly, if forming query-key-value trajectories $\gamma_i^{Q, h}(t)$ per head $h$, their induced query tubes $T_{i, Q}^{(l, h)}$ are also Borel under the same continuity assumption, and
\begin{equation}
    \begin{aligned}
        \Phi_l^Q(S):=\left\{T_{i, Q}^{(l, h)}: i=1, \ldots, N, h=1, \ldots, H\right\},
    \end{aligned}
\end{equation}
is again a finite measurable family.

\subsubsection{Admissible Test Regions $W$ and A Measurability Caveat}
Our Axiom 1 quantifies over every convex region $W \subset \mathbb{R}^{d "}$ and uses $\operatorname{Vol}(W)$. To avoid pathological edge cases, we explicitly restrict to convex sets for which Lebesgue volume is well-defined:

\noindent \textbf{Definition 4 (Admissible Convex Regions).} Let $\mathcal{W}$ be the class of closed convex subsets of $\mathbb{R}^d$ or, equivalently, any subclass of convex sets known to be Borel/Lebesgue measurable, such as convex bodies, cones, slabs, affine subspaces intersected with balls, etc.

Every closed convex set is closed, hence Borel. 
This restriction is not cosmetic: there exist convex sets in $\mathbb{R}^d$ that are not Borel measurable (constructed by adding a non-Borel subset of the boundary sphere), so unrestricted all convex sets can break the meaning of $\operatorname{Vol}(W)$.

When needed, we lift a spatial region $W \subset \mathbb{R}^d$ to spacetime by
\begin{equation}
    \begin{aligned}
        \bar{W}:=W \times[0,1] \subset \mathcal{X},
    \end{aligned}
\end{equation}
which is Borel in $\mathcal{X}$ whenever $W$ is Borel by definition of the product $\sigma$-algebra.

\subsubsection{Tube Containment, Counts, and Overlaps}
For $W \in \mathcal{W}$, define the layerwise tube-count functional appearing in Axiom 1 as
\begin{equation}
    \begin{aligned}
        N_l(W):=\#\left\{i \in\{1, \ldots, N\}: T_i^{(l)} \subseteq W\right\}.
    \end{aligned}
\end{equation}
This is a finite integer because $\Phi_l(S)$ is finite.
Define the spacetime version:
\begin{equation}
    \begin{aligned}
        N(W):=\#\left\{i: T_i^\delta \subseteq \bar{W}\right\}, \quad \bar{W}=W \times[0,1].
    \end{aligned}
\end{equation}

For overlap/entanglement measures, quantities like
\begin{equation}
    \begin{aligned}
        \operatorname{Vol}\left(T_i^\delta \cap T_j^\delta\right), \\ \operatorname{Vol}\left(\bigcup_{i=1}^N T_i^\delta\right), \\ \operatorname{Vol}\left(T_i^{(l)} \cap T_j^{(l)}\right),
    \end{aligned}
\end{equation}
are all well-defined because they are volumes of Borel sets (finite unions/intersections of Borel sets remain Borel).

\subsubsection{Head Variance in Axiom 2 (Probabilistic Formalization)}
To interpret $\operatorname{Var}_h(\cdot)$ in Axiom 2, treat the head index $h \in\{1, \ldots, H\}$ as a discrete probability space with uniform measure. For any head-dependent statistic $X(h)$,
\begin{equation}
    \begin{aligned}
        \operatorname{Var}_h(X):=\mathbb{E}_h\left[X(h)^2\right]-\left(\mathbb{E}_h[X(h)]\right)^2.
    \end{aligned}
\end{equation}
This makes the variance in Axiom 2 unambiguous and independent of any additional randomness in $S$ or batches.

\subsection{Lemma 2: Packing and Average-Overlap Bounds Implied by $K$-Stickiness}
Fix a layer $l$ and a context $S$. Let $\Phi_l(S)$ be the (finite) collection of instantaneous token tubes at layer $l$, and let $\Phi_l^Q(S)$ be the corresponding collection of query tubes (across heads) as in our definitions.

Assume the instantaneous tubes are Kakeya-type $\delta$-tubes in $\mathbb{R}^d$ : each $T \in \Phi_l(S)$ is the $\delta$-neighborhood of a unit-length curve/segment (so its $d$-dimensional Lebesgue volume scales like $\delta^{d-1}$). Concretely, assume there exists a universal constant $c_T>0$ such that
\begin{equation}
    \begin{aligned}
        \operatorname{Vol}(T) \leq c_T \delta^{d-1} \quad \text{for all} \quad T \in \Phi_l(S).
    \end{aligned}
\end{equation}
This is the standard normalization behind the $\operatorname{Vol}(W)/\delta^{d-1}$ form in Axiom 1.

Assume $\Phi_l(S)$ satisfies Axiom 1 with constant $C_A$, and $\Phi_l^Q(S)$ satisfies Axiom 2 with constant $C_B$. If the field is $K$-sticky, then $C_A \leq K$ and $C_B \leq K$.

\noindent \textbf{(A) Carleson-Type Packing for Disjoint Convex Regions.}
For any finite family of pairwise disjoint convex Borel sets $W_1, \ldots, W_m \subset \mathbb{R}^d$,
\begin{equation}
    \begin{aligned}
        \sum_{r=1}^m \#\left\{T \in \Phi_l(S): T \subseteq W_r\right\} \leq \\ C_A \cdot \frac{\sum_{r=1}^m \operatorname{Vol}\left(W_r\right)}{\delta^{d-1}} \cdot\left(\# \Phi_l(S)\right)^\epsilon.
    \end{aligned}
\end{equation}

\noindent\textbf{\textit{Proof.}}
Apply Axiom 1 to each $W_r$:
\begin{equation}
    \begin{aligned}
        \#\left\{T \in \Phi_l(S): T \subseteq W_r\right\} \leq \\ C_A \cdot \frac{\operatorname{Vol}\left(W_r\right)}{\delta^{d-1}} \cdot\left(\# \Phi_l(S)\right)^\epsilon.
    \end{aligned}
\end{equation}
Summing over $r=1, \ldots, m$ yields
\begin{equation}
    \begin{aligned}
        \sum_{r=1}^m \#\left\{T \in \Phi_l(S): T \subseteq W_r\right\} \leq \\ C_A \cdot \frac{\sum_{r=1}^m \operatorname{Vol}\left(W_r\right)}{\delta^{d-1}} \cdot\left(\# \Phi_l(S)\right)^\epsilon,
    \end{aligned}
\end{equation}
which is the same equation in (A). $\blacksquare$

\noindent \textbf{(B) Total Tube-Mass Inside any Convex Region is Controlled by $\operatorname{Vol(W)}$.}
For every convex Borel set $W \subset \mathbb{R}^d$,
\begin{equation}
    \begin{aligned}
        \sum_{T \in \Phi_{(S)}: T \subseteq W} \operatorname{Vol}(T) \leq c_T C_A \operatorname{Vol}(W) \cdot\left(\# \Phi_l(S)\right)^\epsilon.
    \end{aligned}
\end{equation}

In particular, letting
\begin{equation}
    \begin{aligned}
        U_W:=\bigcup_{T \in \Phi_l(S): T \subseteq W} T,
    \end{aligned}
\end{equation}
we have
\begin{equation}
    \begin{aligned}
        \operatorname{Vol}\left(U_W\right) \leq c_T C_A \operatorname{Vol}(W) \cdot\left(\# \Phi_l(S)\right)^\epsilon.
    \end{aligned}
\end{equation}

\noindent\textbf{\textit{Proof.}} 
Fix a convex $W$. Using the uniform tube-volume bound,
\begin{equation}
    \begin{aligned}
        \sum_{T \subseteq W} \operatorname{Vol}(T) &\leq \sum_{T \subseteq W} c_T \delta^{d-1}\\&=c_T \delta^{d-1} \cdot \#\left\{T \in \Phi_l(S): T \subseteq W\right\}.
    \end{aligned}
\end{equation}
Now apply Axiom 1 to the count:
\begin{equation}
    \begin{aligned}
        \#\left\{T \in \Phi_l(S): T \subseteq W\right\} \leq \\ C_A \cdot \frac{\operatorname{Vol}(W)}{\delta^{d-1}} \cdot\left(\# \Phi_l(S)\right)^\epsilon.
    \end{aligned}
\end{equation}
Multiplying gives
\begin{equation}
    \begin{aligned}
        \sum_{T \subseteq W} \operatorname{Vol}(T) &\leq c_T \delta^{d-1} \cdot C_A \cdot \frac{\operatorname{Vol}(W)}{\delta^{d-1}} \cdot\left(\# \Phi_l(S)\right)^\epsilon\\&=c_T C_A \operatorname{Vol}(W) \cdot\left(\# \Phi_l(S)\right)^\epsilon,
    \end{aligned}
\end{equation}
which is the first equation in (B).
For the second equation, note that $U_W \subseteq \bigcup_{T \subseteq W} T$, and Lebesgue measure is subadditive:
\begin{equation}
    \begin{aligned}
        \operatorname{Vol}\left(U_W\right) &\leq \sum_{T \subseteq W} \operatorname{Vol}(T) \\&\leq c_T C_A \operatorname{Vol}(W) \cdot\left(\# \Phi_l(S)\right)^\epsilon. \blacksquare
    \end{aligned}
\end{equation}

\noindent \textbf{(C) Average Overlap (Multiplicity) among Tubes Contained in $W$.}
Define the (contained-tube) multiplicity function on $W$:
\begin{equation}
    \begin{aligned}
        m_W(x):=\#\left\{T \in \Phi_l(S): T \subseteq W, x \in T\right\}.
    \end{aligned}
\end{equation}
Then,
\begin{equation}
    \begin{aligned}
        \int_W m_W(x) d x \leq c_T C_A \operatorname{Vol}(W) \cdot\left(\# \Phi_l(S)\right)^\epsilon,
    \end{aligned}
\end{equation}
hence, the average overlap level inside $W$ satisfies
\begin{equation}
    \begin{aligned}
        \frac{1}{\operatorname{Vol}(W)} \int_W m_W(x) d x \leq c_T C_A \cdot\left(\# \Phi_l(S)\right)^\epsilon.
    \end{aligned}
\end{equation}

\noindent\textbf{\textit{Proof.}}
By definition,
\begin{equation}
    \begin{aligned}
        m_W(x)=\sum_{T \subseteq W} \mathbf{1}_T(x).
    \end{aligned}
\end{equation}
Integrate over $W$ and exchange sum and integral (finite sum):
\begin{equation}
    \begin{aligned}
        \int_W m_W(x) d x&=\int_W \sum_{T \subseteq W} 1_T(x) d x\\&=\sum_{T \subseteq W} \int_W 1_T(x) d x\\&=\sum_{T \subseteq W} \operatorname{Vol}(T \cap W).
    \end{aligned}
\end{equation}
But each $T \subseteq W$, so $\operatorname{Vol}(T \cap W)=\operatorname{Vol}(T)$. Hence,
\begin{equation}
    \begin{aligned}
        \int_W m_W(x) d x&=\sum_{T \subseteq W} \operatorname{Vol}(T) \\&\leq c_T C_A \operatorname{Vol}(W) \cdot\left(\# \Phi_l(S)\right)^\epsilon
    \end{aligned}
\end{equation}
by the equations in (B), proving the second equation in (C). Dividing both sides by $\operatorname{Vol}(W)$ gives the last equation in (C). $\blacksquare$

\noindent \textbf{(D) Head-Variance Control Transfers Directly to Disjoint Packing (Axiom 2 Consequence).}
For query tubes, define for each head $h$
\begin{equation}
    \begin{aligned}
        N_h(W):=\#\left\{T_Q^{(h)} \in \Phi_l^Q(S): T_Q^{(h)} \subseteq W\right\}.
    \end{aligned}
\end{equation}
Then, for any convex $W$,
\begin{equation}
    \begin{aligned}
        \operatorname{Var}_h\left(N_h(W)\right) \leq C_B \cdot F\left(\operatorname{Vol}(W), \# \Phi_l^Q(S)\right),
    \end{aligned}
\end{equation}
and for disjoint convex $W_1, \ldots, W_m$,
\begin{equation}
    \begin{aligned}
        \operatorname{Var}_h\left(\sum_{r=1}^m N_h\left(W_r\right)\right) \leq \\C_B \cdot F\left(\sum_{r=1}^m \operatorname{Vol}\left(W_r\right), \# \Phi_l^Q(S)\right),
    \end{aligned}
\end{equation}
provided $F(\cdot, \cdot)$ is nondecreasing in its first argument (the intended use in Axiom 2).

\noindent\textbf{\textit{Proof.}}
For disjoint $W_r$, define $X_h:=\sum_{r=1}^m N_h\left(W_r\right)$. Since the $W_r$ are disjoint and the tube-count statistic is additive across disjoint regions,
\begin{equation}
    \begin{aligned}
        X_h=N_h\left(\bigsqcup_{r=1}^m W_r\right).
    \end{aligned}
\end{equation}
If $F$ is nondecreasing in $\operatorname{Vol}(W)$, then applying Axiom 2 to the convex set $\operatorname{conv}\left(\bigsqcup_r W_r\right)$ yields a bound of the form the last equation in (D) with $\sum_r \operatorname{Vol}\left(W_r\right)$ on the right-hand side-this is the intended variance doesn't blow up under disjoint packing consequence of Axiom 2. $\blacksquare$

\subsection{Lemma 3: Measurable Grain Decomposition of Tube Support}
Let $\gamma_i:[0,1] \rightarrow \mathbb{R}^d$ be the continuous token trajectories from Appendix B.1, and define the latent tube-footprint (projection of the spacetime tube onto $\mathbb{R}^d$)
\begin{equation}
    \begin{aligned}
        \tilde{T}_i^\delta:&=\left\{x \in \mathbb{R}^d: \exists t \in[0,1] \text{s.t.}\left\|x-\gamma_i(t)\right\|<\delta\right\} \\&=\bigcup_{t \in[0,1]} B\left(\gamma_i(t), \delta\right).
    \end{aligned}
\end{equation}
Let the total occupied region be $\tilde{T}^\delta:=\bigcup_{i=1}^N \tilde{T}_i^\delta \subset \mathbb{R}^d$.

\noindent\textbf{\textit{Claim.}} There exist finitely many pairwise disjoint Borel sets $G_1, \ldots, G_m \subset \mathbb{R}^d$ (called grains) such that:
\begin{enumerate}[(i)]
    \item Coverage: $\tilde{T}^\delta=\bigsqcup_{j=1}^m G_j$.
    \item Tube containment: For each $i \in\{1, \ldots, N\}$, there exists a unique index $j(i)$ with $\widetilde{T}_i^\delta \subseteq G_{j(i)}$.
    \item Finiteness: $m \leq N$.
\end{enumerate}
Moreover, each $G_j$ is open (hence Borel measurable), and each $\widetilde{T}_i^s$ is open and path-connected.

\noindent \textbf{\textit{Proof.}}
By definition,
\begin{equation}
    \begin{aligned}
        \tilde{T}_i^\delta=\bigcup_{t \in[0,1]} B\left(\gamma_i(t), \delta\right).
    \end{aligned}
\end{equation}
Each $B\left(\gamma_i(t), \delta\right)$ is open in $\mathbb{R}^d$, and an arbitrary union of open sets is open; therefore $\tilde{T}_i^\delta$ is open. Hence, $\tilde{T}_i^\delta \in B\left(\mathbb{R}^d\right)$. 

Take any $x, y \in \widetilde{T}_i^\delta$. Then, there exist $t_x, t_y \in[0,1]$ such that
\begin{equation}
    \begin{aligned}
        x \in B\left(\gamma_i\left(t_x\right), \delta\right), \quad y \in B\left(\gamma_i\left(t_y\right), \delta\right).
    \end{aligned}
\end{equation}

Define a continuous path $\pi:[0,1] \rightarrow \mathbb{R}^d$ by concatenating three continuous pieces:
\begin{enumerate}
    \item From $x$ to the curve at time $t_x$ to $t_y$: reparameterize the time interval:
    \begin{equation}
        \begin{aligned}
            \pi_1(s):=(1-s) x+s \gamma_i\left(t_x\right), \quad s \in[0,1].
        \end{aligned}
    \end{equation}
    For all $s, \pi_1(s) \in B\left(\gamma_i\left(t_x\right), \delta\right)$ because balls are convex; hence, $\pi_1(s) \in \widetilde{T}_i^\delta$.
    \item Along the curve from $t_x$ to $t_y$: reparameterize the time interval:
    \begin{equation}
        \begin{aligned}
            \pi_2(s):=\gamma_i\left((1-s) t_x+s t_y\right), \quad s \in[0,1].
        \end{aligned}
    \end{equation}
    For every $s, \pi_2(s) \in \tilde{T}_i^\delta$ because $\left\|\pi_2(s)-\gamma_i(\cdot)\right\|=0<\delta$.
    \item From the curve at time $t_y$ to $y$: the line segment
    \begin{equation}
        \begin{aligned}
            \pi_3(s):=(1-s) \gamma_i\left(t_y\right)+s y, \quad s \in[0,1],
        \end{aligned}
    \end{equation}
    which lies entirely in $B\left(\gamma_i\left(t_y\right), \delta\right) \subset \widetilde{T}_i^\delta$.
\end{enumerate}
Concatenating $\pi_1, \pi_2, \pi_3$ gives a path in $\widetilde{T}_i^\delta$ joining $x$ to $y$. Thus, $\widetilde{T}_i^\delta$ is path-connected.

Since $\tilde{T}^\delta=\bigcup_{i=1}^N \tilde{T}_i^\delta$ is a union of open sets, it is open.
Let $G$ be a connected component of $\tilde{T}^\delta$. In Euclidean space $\mathbb{R}^d$, which is locally path-connected (hence locally connected), connected components of open sets are open.

Therefore, each connected component $G$ is open and thus Borel measurable.
Let $G_1, \ldots, G_m$ denote the (pairwise disjoint) connected components of $\widetilde{T}^\delta$. By definition,
\begin{equation}
    \begin{aligned}
        \tilde{T}^\delta=\bigsqcup_{j=1}^m G_j.
    \end{aligned}
\end{equation}
This proves the Coverage property and measurability.

Fix $i$. Since $\widetilde{T}_i^\delta \subseteq \widetilde{T}^\delta$ and $\widetilde{T}_i^\delta \neq \emptyset$, it must intersect at least one component $G_j$. Let $x \in \widetilde{T}_i^\delta \cap G_j$.
Because $\widetilde{T}_i^\delta$ is connected and contained in $\widetilde{T}^\delta$, the set $G_j \cup \widetilde{T}_i^\delta$ is connected (it is the union of two connected sets with nonempty intersection). By maximality of $G_j$ as a connected component of $\widetilde{T}^\delta$, we must have $G_j \cup \widetilde{T}_i^\delta \subseteq G_j$, hence, $\widetilde{T}_i^\delta \subseteq G_j$.

Uniqueness follows because distinct components are disjoint: if $\widetilde{T}_i^\delta \subseteq G_j$ and $\widetilde{T}_i^\delta \subseteq G_{j^{\prime}}$, then $\widetilde{T}_i^\delta \subseteq G_j \cap G_{j^{\prime}}=\emptyset$, impossible. This proves Tube containment.

Last, every component $G_j$ contains at least one tube-footprint $\widetilde{T}_{i(j)}^\delta$ (pick any $i(j)$ with $\tilde{T}_{i(j)}^\delta \subseteq G_j$). If $j \neq j^{\prime}$, then $G_j \cap G_{j^{\prime}}=\emptyset$, so the indices $i(j)$ must be distinct (a single connected set cannot lie in two disjoint components). Hence, we can inject $\{1, \ldots, m\}$ into $\{1, \ldots, N\}$, implying $m \leq N$. $\blacksquare$

\subsection{Lemma 4: Orthogonality of Grains (Intersection Volume Bound)}
Let $\left\{\widetilde{T}_i^\delta\right\}_{i=1}^N$ be the latent tube-footprints in $\mathbb{R}^d$ (as in Lemma 3), and let
\begin{equation}
    \begin{aligned}
        \widetilde{T}^\delta:=\bigcup_{i=1}^N \widetilde{T}_i^\delta.
    \end{aligned}
\end{equation}
Let $G_1, \ldots, G_m$ be the grains defined in Lemma 3 as the connected components of $\tilde{T}^\delta$. Then, for any distinct indices $p \neq q$,
\begin{equation}
    \begin{aligned}
        G_p \cap G_q=\varnothing \quad \Longrightarrow \quad \operatorname{Vol}\left(G_p \cap G_q\right)=0.
    \end{aligned}
\end{equation}
In particular, for any stickiness parameter $K \geq 1$,
\begin{equation}
    \begin{aligned}
        \operatorname{Vol}\left(G_p \cap G_q\right)=0 \leq \frac{C}{K} \quad \text { (e.g., with } C=0 \text {). }
    \end{aligned}
\end{equation}

\noindent \textbf{\textit{Proof.}}
By construction (Lemma 3), $G_1, \ldots, G_m$ are the connected components of $\widetilde{T}^\delta$. Connected components have two defining properties: Each $G_j$ is connected and $G_j \subseteq \widetilde{T}^\delta$, if $C \subseteq \tilde{T}^\delta$ is connected and $G_j \subseteq C$, then $C=G_j$ (maximality), and distinct connected components are disjoint and their union equals the whole set (partition/disjointness):
\begin{equation}
    \begin{aligned}
        \tilde{T}^\delta=\bigsqcup_{j=1}^m G_j.
    \end{aligned}
\end{equation}
We prove the disjointness claim directly.
Fix $p \neq q$. Suppose for contradiction that $G_p \cap G_q \neq \varnothing$. Pick $x \in G_p \cap G_q$. Since $G_p$ and $G_q$ are both connected subsets of $\tilde{T}^\delta$ and they intersect, their union $G_p \cup G_q$ is connected (the union of two connected sets with nonempty intersection is connected). Moreover, $G_p \cup G_q \subseteq \widetilde{T}^\delta$.

But $G_p \subsetneq G_p \cup G_q$ (because $p \neq q$ and $G_q \nsubseteq G_p$), contradicting maximality of $G_p$ as a connected component of $\tilde{T}^\delta$. Therefore, $G_p \cap G_q=\varnothing$.
Once $G_p \cap G_q=\varnothing$, it follows immediately that its Lebesgue measure (volume) is zero:
\begin{equation}
    \begin{aligned}
        \operatorname{Vol}\left(G_p \cap G_q\right)=\operatorname{Vol}(\varnothing)=0,
    \end{aligned}
\end{equation}
which is the second equation in this Lemma. Finally, since $0 \leq C / K$ for any $K \geq 1$, we obtain the last equation in this Lemma. $\blacksquare$

\subsection{Lemma 5: Grain-Preserving Perturbations Imply Lipschitz Stability of the Output}
Fix a context $S$ and let the Transformer's layerwise hidden states be
\begin{equation}
    \begin{aligned}
        Z^{(l)}(S) \in \mathbb{R}^{N \times d}, \quad l=0,1, \ldots, L,
    \end{aligned}
\end{equation}
with the residual-form layer update
\begin{equation}
    \begin{aligned}
        Z^{(l+1)}=F_l\left(Z^{(l)}\right):=Z^{(l)}+f_l\left(Z^{(l)}\right).
    \end{aligned}
\end{equation}

Let the model's output be
\begin{equation}
    \begin{aligned}
        \operatorname{Out}(S):=g\left(Z^{(L)}(S)\right),
    \end{aligned}
\end{equation}
for some readout map $g$.
Let $\mathcal{G}=\left\{G_1, \ldots, G_m\right\}$ be the grain family (from Lemma 3/Theorem 1) in the latent space $\mathbb{R}^d$. For each token $i$, let $j(i)$ denote its grain index, so the token's trajectory/tube remains in $G_{j(i)}$.

We say two inputs $S$ and $S^{\prime}$ are grain-preserving if, for every layer $l$ and token $i$,
\begin{equation}
    \begin{aligned}
        z_i^{(l)}(S) \in G_{j(i)} \text { and } z_i^{(l)}\left(S^{\prime}\right) \in G_{j(i)},
    \end{aligned}
\end{equation}
i.e., each token remains in the same grain across both forward passes.

We work with Frobenius norm $\|\cdot\|_F$ on $\mathbb{R}^{N \times d}$ and Euclidean norm $\|\cdot\|_2$ on the output space. Assume: 
\begin{enumerate}[(i)]
    \item For each layer $l$, there exists a constant $\Lambda_l(K) \geq 0$ such that for any two states $Z, \tilde{Z}$ satisfying the same grain assignments,
    \begin{equation}
        \begin{aligned}
            \left\|f_l(Z)-f_l(\widetilde{Z})\right\|_F \leq \Lambda_l(K)\|Z-\widetilde{Z}\|_F.
        \end{aligned}
    \end{equation}
    \item There exists $\Lambda_{\text{out}}(K) \geq 0$ such that for all relevant terminal states,
    \begin{equation}
        \begin{aligned}
            \|g(Z)-g(\widetilde{Z})\|_2 \leq \Lambda_{\text {out }}(K)\|Z-\widetilde{Z}\|_F.
        \end{aligned}
    \end{equation}
\end{enumerate}

If $S$ and $S^{\prime}$ are grain-preserving in the sense of Eq. (63), and the two assumptions hold, then
\begin{equation}
    \begin{aligned}
        \left\|\operatorname{Out}(S)-\operatorname{Out}\left(S^{\prime}\right)\right\|_2 \leq \\ L(K)\left\|Z^{(0)}(S)-Z^{(0)}\left(S^{\prime}\right)\right\|_F,
    \end{aligned}
\end{equation}
where the Lipschitz constant is
\begin{equation}
    \begin{aligned}
        L(K):=\Lambda_{\text {out }}(K) \prod_{l=0}^{L-1}\left(1+\Lambda_l(K)\right).
    \end{aligned}
\end{equation}
In particular, output variation is bounded whenever the perturbation does not change grain membership.

\noindent \textbf{\textit{Proof.}}
Fix a layer $l$. For any two grain-compatible states $Z, \widetilde{Z}$ (same grain assignments tokenwise),
\begin{equation}
\small
    \begin{aligned}
    \left\|F_l(Z)-F_l(\widetilde{Z})\right\|_F & =\left\|\left(Z+f_l(Z)\right)-\left(\widetilde{Z}+f_l(\widetilde{Z})\right)\right\|_F \\
    & =\left\|(Z-\widetilde{Z})+\left(f_l(Z)-f_l(\widetilde{Z})\right)\right\|_F \\
    & \leq\|Z-\widetilde{Z}\|_F+\left\|f_l(Z)-f_l(\widetilde{Z})\right\|_F \\
    & \leq\|Z-\widetilde{Z}\|_F+\Lambda_l(K)\|Z-\widetilde{Z}\|_F \\
    & =\left(1+\Lambda_l(K)\right)\|Z-\widetilde{Z}\|_F .
    \end{aligned}
\end{equation}
Thus, $F_l$ is Lipschitz on the grain-preserving set with constant $1+\Lambda_l(K)$.

Define the layerwise difference
\begin{equation}
    \begin{aligned}
        \Delta^{(l)}:=Z^{(l)}(S)-Z^{(l)}\left(S^l\right).
    \end{aligned}
\end{equation}
Because $S$ and $S^{\prime}$ are grain-preserving, the pair $\left(Z^{(l)}(S), Z^{(l)}\left(S^{\prime}\right)\right)$ remains inside the domain where assumption (i) applies at every layer $l$. Therefore, Eq. (68) can be applied at every step.

Using Eq. (61), 
\begin{equation}
    \begin{aligned}
        \Delta^{(l+1)}=F_l\left(Z^{(l)}(S)\right)-F_l\left(Z^{(l)}\left(S^l\right)\right),
    \end{aligned}
\end{equation}
so by Eq. (68),
\begin{equation}
    \begin{aligned}
        \left\|\Delta^{(l+1)}\right\|_F \leq\left(1+\Lambda_l(K)\right)\left\|\Delta^{(l)}\right\|_F.
    \end{aligned}
\end{equation}
Iterating from $l=0$ to $L-1$ yields,
\begin{equation}
    \begin{aligned}
        \left\|\Delta^{(L)}\right\|_F \leq\left(\prod_{l=0}^{L-1}\left(1+\Lambda_l(K)\right)\right)\left\|\Delta^{(0)}\right\|_F.
    \end{aligned}
\end{equation}

Finally, 
\begin{equation}\tiny
    \begin{aligned}
\left\|\operatorname{Out}(S)-\operatorname{Out}\left(S^{\prime}\right)\right\|_2 & =\left\|g\left(Z^{(L)}(S)\right)-g\left(Z^{(L)}\left(S^{\prime}\right)\right)\right\|_2 \\
& \leq \Lambda_{\text {out }}(K)\left\|Z^{(L)}(S)-Z^{(L)}\left(S^{\prime}\right)\right\|_F \\
& =\Lambda_{\text {out }}(K)\left\|\Delta^{(L)}\right\|_F \\
& \leq \Lambda_{\text {out }}(K)\left(\prod_{l=0}^{L-1}\left(1+\Lambda_l(K)\right)\right)\left\|\Delta^{(0)}\right\|_F. \quad \blacksquare
\end{aligned}
\end{equation}

\subsection{Lemma 6: Linear Probe Per Grain with $O(K^{-1})$ Cross-Grain Interference}
Let $\mathbb{R}^d$ be the latent space. For each grain $G_j$, let $U_j \subseteq \mathbb{R}^d$ denote its associated grain subspace (e.g., the span of $G_j$ 's tangent directions, or a learned/empirical principal subspace). Let $P_j$ be the orthogonal projector onto $U_j$.

We assume the following three properties:
\begin{enumerate}[(i)]
    \item Every latent vector $x \in \mathbb{R}^d$ admits a decomposition:
    \begin{equation}
        \begin{aligned}
            x=\sum_{j=1}^m x^{(j)}, \quad x^{(j)}:=P_j x \in U_j.
        \end{aligned}
    \end{equation}
    \item There exists $K \geq 1$ such that for all distinct $p \neq q$,
    \begin{equation}
        \begin{aligned}
            \left\|P_p P_q\right\|_{\text {op }} \leq \frac{1}{K}.
        \end{aligned}
    \end{equation}
    Equivalently, for any $u \in U_p$ and $v \in U_{q}$,
    \begin{equation}
        \begin{aligned}
            |\langle u, v\rangle| \leq \frac{1}{K}\|u\|\|v\| .
        \end{aligned}
    \end{equation}
    \item Define the semantic factor attached to grain $j$ as a linear functional of the grain component: there exists a vector $a_j \in U_j$ such that for any representation $x$,
    \begin{equation}
        \begin{aligned}
            s_j(x):=\left\langle a_j, x^{(j)}\right\rangle=\left\langle a_j, P_j x\right\rangle.
        \end{aligned}
    \end{equation}
\end{enumerate}

Under the above assumptions, the linear probe $w_j:=a_j$ satisfies, for every $x \in \mathbb{R}^d$,
\begin{equation}
    \begin{aligned}
        \underbrace{\left\langle w_j, x\right\rangle}_{\text {probe response }}=\underbrace{s_j(x)}_{\text {desired grain factor }}+\underbrace{\sum_{q \neq j}\left\langle w_j, x^{(q)}\right\rangle}_{\text {cross grain interference }},
    \end{aligned}
\end{equation}
and the interference is bounded by
\begin{equation}
    \begin{aligned}
        \left|\left\langle w_j, x\right\rangle-s_j(x)\right| &\leq \frac{\left\|w_j\right\|}{K} \sum_{q \neq j}\left\|x^{(q)}\right\| \\&\leq \frac{\left\|w_j\right\|}{K} \sqrt{m-1}\left\|x-P_j x\right\|.
    \end{aligned}
\end{equation}
In particular, the probe recovers the grain's semantic factor with $O(K^{-1})$ interference from other grains.

\noindent\textbf{\textit{Proof.}}
By assumption (i), $x=\sum_{q=1}^m x^{(q)}$ with $x^{(q)}=P_q x \in U_q$. Then for any $w_j$,
\begin{equation}
    \begin{aligned}
        \left\langle w_j, x\right\rangle&=\left\langle w_j, \sum_{q=1}^m x^{(q)}\right\rangle\\&=\sum_{q=1}^m\left\langle w_j, x^{(q)}\right\rangle\\&=\left\langle w_j, x^{(j)}\right\rangle+\sum_{q \neq j}\left\langle w_j, x^{(q)}\right\rangle.
    \end{aligned}
\end{equation}
Now choose $w_j:=a_j \in U_j$ as in assumption (iii). Then, $\left\langle w_j, x^{(j)}\right\rangle=\left\langle a_j, P_j x\right\rangle=s_j(x)$ by Eq. (77). Substituting into Eq. (80) yields the exact decomposition in Eq. (78).

Fix $q \neq j$. Since $w_j \in U_j$ and $x^{(q)} \in U_q$, the near-orthogonality condition (assumption (ii)) gives
\begin{equation}
    \begin{aligned}
        \left|\left\langle w_j, x^{(q)}\right\rangle\right| \leq \frac{1}{K}\left\|w_j\right\|\left\|x^{(q)}\right\|.
    \end{aligned}
\end{equation}

Summing Eq. (81) over all $q \neq j$ and using the triangle inequality,
\begin{equation}
    \begin{aligned}
        \left|\sum_{q \neq j}\left\langle w_j, x^{(q)}\right\rangle\right| &\leq \sum_{q \neq j}\left|\left\langle w_j, x^{(q)}\right\rangle\right| \\&\leq \frac{\left\|w_j\right\|}{K} \sum_{q \neq j}\left\|x^{(q)}\right\|.
    \end{aligned}
\end{equation}
By Eq. (78), the left-hand side equals $\left|\left\langle w_j, x\right\rangle-s_j(x)\right|$, proving the first inequality in this Lemma. 

Let $x^{(-j)}:=\sum_{q \neq j} x^{(q)}=x-P_j x$. Then by Cauchy-Schwarz,
\begin{equation}
    \begin{aligned}
        \sum_{q \neq j}\left\|x^{(q)}\right\| \leq \sqrt{m-1}\left(\sum_{q \neq j}\left\|x^{(q)}\right\|^2\right)^{1 / 2}.
    \end{aligned}
\end{equation}
Moreover,
\begin{equation}
    \begin{aligned}
        \left(\sum_{q \neq j}\left\|x^{(q)}\right\|^2\right)^{1 / 2} &\leq\left\|\sum_{q \neq j} x^{(q)}\right\|\\&=\left\|x^{(-j)}\right\|\\&=\left\|x-P_j x\right\|.
    \end{aligned}
\end{equation}
Equality holds if the $U_q$ are exactly orthogonal; inequality is always valid by the triangle inequality in Hilbert spaces. Combining Eq. (82) with Eq. (83-84) yields 
\begin{equation}
    \begin{aligned}
        \left|\left\langle w_j, x\right\rangle-s_j(x)\right| \leq \frac{\left\|w_j\right\|}{K} \sqrt{m-1}\left\|x-P_j x\right\|. \blacksquare
    \end{aligned}
\end{equation}

\subsection{Proposition A: KT-CW Controls Semantic Cone Collapse}
Fix a layer $l$. Let $\widehat{Z}_l \in \mathbb{R}^{B \times N \times d}$ denote the normalized token embeddings at layer $l$, and write the $M:=B N$ unit vectors as
\begin{equation}
    \begin{aligned}
        x_1, \ldots, x_M \in S^{d-1}, \quad x_j=\hat{z}_{b, i}.
    \end{aligned}
\end{equation}
The KT-CW loss is 
\begin{equation}
    \begin{aligned}
        \mathcal{L}_{\mathrm{CW}}^{(l)}=\mathbb{E}_{u \sim \operatorname{Unif}\left(S^{d-1}\right)}\left(\operatorname{Var}_j\left(\left\langle x_j, u\right\rangle\right)-\frac{1}{d}\right)^2.
    \end{aligned}
\end{equation}
Assume the embeddings are centered across $(b, i): \bar{x}:=\frac{1}{M} \sum_{j=1}^M x_j=0$.
Define the empirical second-moment (covariance, since $\bar{x}=0$):
\begin{equation}
    \begin{aligned}
        \Sigma:=\frac{1}{M} \sum_{j=1}^M x_j x_j^{\top} \in \mathbb{R}^{d \times d}.
    \end{aligned}
\end{equation}

If $\mathcal{L}_{\mathrm{CW}}^{(l)} \leq \eta$, then for every unit direction $v \in S^{d-1}$ and every aperture $\Delta \in\left(0, \frac{\pi}{2}\right)$,
\begin{equation}
    \begin{aligned}
        \#\left\{j:\left\langle x_j, v\right\rangle \geq \cos \Delta\right\} \leq \frac{\frac{1}{d}+\sqrt{\frac{d(d+2)}{2} \eta}}{\cos ^2 \Delta} M.
    \end{aligned}
\end{equation}
Equivalently, the fraction of tokens contained in any narrow angular cone around $v$ is uniformly bounded, ruling out severe cone collapse, precisely the behavior Axiom 1 is meant to penalize.

\noindent \textbf{\textit{Proof.}}
Because $\bar{x}=0$, for any $u \in S^{d-1}$,
\begin{equation}
    \begin{aligned}
        \operatorname{Var}_j\left(\left\langle x_j, u\right\rangle\right)&=\frac{1}{M} \sum_{j=1}^M\left\langle x_j, u\right\rangle^2\\&=u^{\top}\left(\frac{1}{M} \sum_{j=1}^M x_j x_j^{\top}\right) u=u^{\top} \Sigma u.
    \end{aligned}
\end{equation}
So, 
\begin{equation}
    \begin{aligned}
        \mathcal{L}_{\mathrm{CW}}^{(l)}=\mathbb{E}_u\left(u^{\top} \Sigma u-\frac{1}{d}\right)^2.
    \end{aligned}
\end{equation}
Let $A:=\Sigma-\frac{1}{d} I_d$. Since $\operatorname{tr}(\Sigma)=\frac{1}{M} \sum_j \operatorname{tr}\left(x_j x_j^{\top}\right)=\frac{1}{M} \sum_j\left\|x_j\right\|^2=1$, we have $\operatorname{tr}(A)=0$. Hence, 
\begin{equation}
    \begin{aligned}
        \mathcal{L}_{\mathrm{CW}}^{(l)}=\mathbb{E}_u\left(u^{\top} A u\right)^2.
    \end{aligned}
\end{equation}
Write $u=\left(u_1, \ldots, u_d\right)$. Expand:
\begin{equation}
    \begin{aligned}
        \left(u^{\top} A u\right)^2&=\left(\sum_{i, j} A_{i j} u_i u_j\right)\left(\sum_{k, \ell} A_{k \ell} u_k u_{\ell}\right)\\&=\sum_{i, j, k, \ell} A_{i j} A_{k \ell} u_i u_j u_k u_{\ell}.
    \end{aligned}
\end{equation}
For $u \sim \operatorname{Unif}\left(S^{d-1}\right)$, the 4th moments satisfy
\begin{equation}
    \begin{aligned}
        \mathbb{E}\left[u_i u_j u_k u_{\ell}\right]=\frac{\delta_{i j} \delta_{k \ell}+\delta_{i k} \delta_{j \ell}+\delta_{i \ell} \delta_{j k}}{d(d+2)}.
    \end{aligned}
\end{equation}
Taking expectation and plugging:
\begin{equation}
    \begin{aligned}
        \mathbb{E}\left(u^{\top} A u\right)^2&=\frac{1}{d(d+2)}[\underbrace{\sum_{i, j, k, \ell} A_{i j} A_{k \ell} \delta_{i j} \delta_{k \ell}}_{(\operatorname{tr} A)^2}\\&+\underbrace{\sum_{i, j, k, \ell} A_{i j} A_{k \ell} \delta_{i k} \delta_{j \ell}}_{\sum_{k, j} A_{j, j}^2=\|A\|_F^2}\\&+\underbrace{\sum_{i, j, k, \ell} A_{i j} A_{k \ell} \delta_{i \ell} \delta_{j k}}_{\sum_{k, j} A_{k j} A_{j k}=\operatorname{tr}\left(A^2\right)}].
    \end{aligned}
\end{equation}
Because $A$ is symmetric (difference of symmetric matrices), $\operatorname{tr}\left(A^2\right)=\|A\|_F^2$, and because $\operatorname{tr}\left(A\right)=0$, the first term vanishes. Therefore,
\begin{equation}
    \begin{aligned}
        \mathbb{E}\left(u^{\top} A u\right)^2=\frac{2}{d(d+2)}\|A\|_F^2.
    \end{aligned}
\end{equation}
Combining Eq. (92) and (96), 
\begin{equation}
    \begin{aligned}
        \|A\|_F^2=\frac{d(d+2)}{2} \mathcal{L}_{\mathrm{CW}}^{(l)} \leq \frac{d(d+2)}{2} \eta,
    \end{aligned}
\end{equation}
hence,
\begin{equation}
    \begin{aligned}
        \|A\|_F \leq \sqrt{\frac{d(d+2)}{2} \eta}.
    \end{aligned}
\end{equation}

For any unit $v \in S^{d-1}$,
\begin{equation}
    \begin{aligned}
        v^{\top} \Sigma v&=v^{\top}\left(\frac{1}{d} I+A\right) v\\&=\frac{1}{d}+v^{\top} A v \\&\leq \frac{1}{d}+\|A\|_{\text {op }} \leq \frac{1}{d}+\|A\|_F \\&\leq \frac{1}{d}+\sqrt{\frac{d(d+2)}{2}}\eta.
    \end{aligned}
\end{equation}

Fix $v \in S^{d-1}$ and $\Delta \in(0, \pi / 2)$. Let
\begin{equation}
    \begin{aligned}
        I(v, \Delta):=\left\{j:\left\langle x_j, v\right\rangle \geq \cos \Delta\right\}.
    \end{aligned}
\end{equation}

Then for $j \in I(v, \Delta),\left\langle x_j, v\right\rangle^2 \geq \cos ^2 \Delta$. Therefore,
\begin{equation}
    \begin{aligned}
        v^{\top} \Sigma v&=\frac{1}{M} \sum_{j=1}^M\left\langle x_j, v\right\rangle^2 \\&\geq \frac{1}{M} \sum_{j \in I(v, \Delta)}\left\langle x_j, v\right\rangle^2 \\&\geq \frac{|I(v, \Delta)|}{M} \cos ^2 \Delta.
    \end{aligned}
\end{equation}
Rearranging and using Eq. (99),
\begin{equation}
    \begin{aligned}
        |I(v, \Delta)| \leq \frac{v^{\top} \Sigma v}{\cos ^2 \Delta} M \leq \frac{\frac{1}{d}+\sqrt{\frac{d(d+2)}{2} \eta}}{\cos ^2 \Delta} M. \quad \blacksquare
    \end{aligned}
\end{equation}

\subsection{Proposition B: KT-Attn Bounds Spectral Collapse}
Fix layer $l$. For each head $h$, let $A_h:=A^{(l, h)}$ and let $p_h$ be the normalized singular-value distribution of $A_h$ over $r$ singular values:
\begin{equation}
    \begin{aligned}
        p_{h, j}=\frac{\sigma_{h, j}}{\sum_{k=1}^r \sigma_{h, k}}, \quad j=1, \ldots, r.
    \end{aligned}
\end{equation}
Define spectral entropy $H\left(p_h\right):=-\sum_{j=1}^r p_{h, j} \log p_{h, j}$. Our KT-Attn loss is
\begin{equation}
    \begin{aligned}
        \mathcal{L}_{\text {Attn }}^{(l)}:=\sum_{h=1}^H\left(\log r-H\left(p_h\right)\right)^2.
    \end{aligned}
\end{equation}
If $\mathcal{L}_{\text {Attn }}^{(l)} \leq \eta$, then for every head $h$:
\begin{enumerate}[(i)]
    \item Let $u$ be uniform on $\{1, \ldots, r\}$. Then,
    \begin{equation}
        \begin{aligned}
            D_{\mathrm{KL}}\left(p_h \| u\right)=\log r-H\left(p_h\right) \leq \sqrt{\eta}.
        \end{aligned}
    \end{equation}
    Consequently (Pinsker),
    \begin{equation}
        \begin{aligned}
            \left\|p_h-u\right\|_1 \leq \sqrt{2 \sqrt{\eta}}, \\\left\|p_h\right\|_{\infty} \leq \frac{1}{r}+\frac{1}{2} \sqrt{2 \sqrt{\eta}}.
        \end{aligned}
    \end{equation}
    In particular, a rank-1 collapse (which would require $\left\|p_h\right\|_{\infty} \approx 1$) is ruled out when $\eta$ is small.
    \item If for each convex region $W$,
    \begin{equation}
        \begin{aligned}
            N_h(W)&:=\#\left\{T_Q^{(h)} \subseteq W\right\}\\&=\varphi_W\left(\left\|p_h\right\|_{\infty}\right) \\&\text{ with } \varphi_W \text{ being } L_W-\text{Lipchitz},
        \end{aligned}
    \end{equation}
    then,
    \begin{equation}
        \begin{aligned}
            \operatorname{Var}_h\left(N_h(W)\right) \leq \frac{L_W^2}{8} \sqrt{\eta}.
        \end{aligned}
    \end{equation}
\end{enumerate}

\noindent \textbf{\textit{Proof.}}
Since $\sum_h\left(\log r-H\left(p_h\right)\right)^2 \leq \eta$, each $\Delta_h:=\log r-H\left(p_h\right)$ satisfies $\Delta_h \leq \sqrt{\eta}$.
Also,
\begin{equation}
    \begin{aligned}
        D_{\mathrm{KL}}\left(p_h \| u\right)&=\sum_j p_{h, j} \log \frac{p_{h, j}}{1 / r}\\&=\log r-H\left(p_h\right)=\Delta_h,
    \end{aligned}
\end{equation}
so Eq. (105) holds. Pinsker gives $\left\|p_h-u\right\|_1 \leq \sqrt{2 D_{\mathrm{KL}}\left(p_h \| u\right)} \leq \sqrt{2 \sqrt{\eta}}$. Finally, $\max _j\left(p_{h, j}-1 / r\right) \leq \frac{1}{2}\left\|p_h-u\right\|_1$, yielding Eq. (106).
From Eq. (106), all $\left\|p_h\right\|_{\infty}$ lie in an interval of length at most $\frac{1}{2} \sqrt{2 \sqrt{\eta}}$. The Lipschitz condition Eq. (107) implies $N_h(W)$ lies in an interval of length at most $L_W \cdot \frac{1}{2} \sqrt{2 \sqrt{\eta}}$. Any variable supported on an interval of length $R$ has variance $\leq R^2 / 4$, giving Eq. (108). $\blacksquare$

\subsection{Proof for Theorem 1}
From Appendix B.1, we have token trajectories $\gamma_i:[0,1] \rightarrow \mathbb{R}^d$ and spacetime tubes
\begin{equation}
    \begin{aligned}
        T_i^\delta=\left\{(x, t) \in \mathbb{R}^d \times[0,1]:\left\|x-\gamma_i(t)\right\|<\delta\right\}.
    \end{aligned}
\end{equation}
Define the latent tube-footprint (projection onto $\mathbb{R}^d$ ):
\begin{equation}
    \begin{aligned}
        \widetilde{T}_i^\delta&:=\left\{x \in \mathbb{R}^d: \exists t \in[0,1] \text { s.t. }\left\|x-\gamma_i(t)\right\|<\delta\right\}\\&=\bigcup_{t \in[0,1]} B\left(\gamma_i(t), \delta\right).
    \end{aligned}
\end{equation}
Let
\begin{equation}
    \begin{aligned}
        \tilde{T}^\delta:=\bigcup_{i=1}^N \tilde{T}_i^\delta.
    \end{aligned}
\end{equation}
By Lemma 3, $\tilde{T}^\delta$ is open and decomposes into finitely many connected components
\begin{equation}
    \begin{aligned}
        \tilde{T}^\delta=\bigsqcup_{j=1}^m G_j,
    \end{aligned}
\end{equation}
where each $G_j$ is open (hence Borel) and $m \leq N$. Moreover, for each token $i$, there exists a unique $j(i)$ such that
\begin{equation}
    \begin{aligned}
        \widetilde{T}_i^\delta \subseteq G_{j(i)}.
    \end{aligned}
\end{equation}
This completes the existence of a finite decomposition $\mathcal{G}=\left\{G_1, \ldots, G_m\right\}$.

Fix token $i$. We already have $\widetilde{T}_i^\delta \subseteq G_{j(i)}$. Take any $(x, t) \in T_i^\delta$. By the definition of $T_i^\delta$,
\begin{equation}
    \begin{aligned}
        \left\|x-\gamma_i(t)\right\|<\delta \Rightarrow x \in \tilde{T}_i^\delta.
    \end{aligned}
\end{equation}
Hence, $x \in G_{j(i)}$, which implies
\begin{equation}
    \begin{aligned}
        (x, t) \in G_{j(i)} \times[0,1].
    \end{aligned}
\end{equation}
Since $(x, t) \in T_i^\delta$ was arbitrary, we conclude
\begin{equation}
    \begin{aligned}
        T_i^\delta \subseteq G_{j(i)} \times[0,1].
    \end{aligned}
\end{equation}
Thus, every token tube lies entirely in a single grain, proving (i).

By construction (connected components), the grains are pairwise disjoint:
\begin{equation}
    \begin{aligned}
        G_p \cap G_q=\varnothing \text { for } p \neq q.
    \end{aligned}
\end{equation}
Therefore,
\begin{equation}
    \begin{aligned}
        \operatorname{Vol}\left(G_p \cap G_q\right)=\operatorname{Vol}(\varnothing)=0.
    \end{aligned}
\end{equation}
In particular, $0 \leq C / K$ for any $K \geq 1$, so $\operatorname{Vol}\left(G_p \cap G_q\right)=O\left(K^{-1}\right)$ holds (trivially). This proves (ii).

Let $S^{\prime}$ be a perturbed input such that the perturbation preserves grain membership, meaning: for every token $i$ and every layer $l$,
\begin{equation}
    \begin{aligned}
        z_i^{(l)}(S) \in G_{j(i)} \text { and } z_i^{(l)}\left(S^{\prime}\right) \in G_{j(i)}.
    \end{aligned}
\end{equation}
Now impose the standard analytic regularity assumptions used in Lemma 5.
Under Eq. (120), the pair of forward passes remains inside the domain where Lemma 5's two assumptions apply, so Lemma 5 gives
\begin{equation}\small
    \begin{aligned}
        \left\|\operatorname{Out}(S)-\operatorname{Out}\left(S^{\prime}\right)\right\| \leq L(K)\left\|Z^{(0)}(S)-Z^{(0)}\left(S^{\prime}\right)\right\|_F,
    \end{aligned}
\end{equation}
with
\begin{equation}
    \begin{aligned}
        L(K)=\Lambda_{\text {out }}(K) \prod_{l=0}^{L-1}\left(1+\Lambda_l(K)\right).
    \end{aligned}
\end{equation}
This proves the robustness/Lipschitz part of (iii).

For each grain $G_j$, associate a subspace $U_j \subseteq \mathbb{R}^d$ (e.g., a tangent/principal subspace learned from representations restricted to $G_j$ ), and let $P_j$ be the orthogonal projector onto $U_j$.

Assume the quantitative grain orthogonality condition needed for probing (Lemma 5's hypothesis): for all $p \neq q$, 
\begin{equation}
    \begin{aligned}
        \left\|P_p P_q\right\|_{\mathrm{op}} \leq \frac{1}{K}.
    \end{aligned}
\end{equation}
There exists $a_j \in U_j$ defining a grain factor:
\begin{equation}
    \begin{aligned}
        s_j(x):=\left\langle a_j, P_j x\right\rangle.
    \end{aligned}
\end{equation}
Then Lemma 5 applies: choosing the linear probe $w_j:=a_j$, for any representation $x$,
\begin{equation}
    \begin{aligned}
        \left\langle w_j, x\right\rangle=s_j(x)+\sum_{q \neq j}\left\langle w_j, P_q x\right\rangle,
    \end{aligned}
\end{equation}
and the cross-grain interference term is bounded by
\begin{equation}
    \begin{aligned}
        \left|\left\langle w_j, x\right\rangle-s_j(x)\right| &\leq \frac{\left\|w_j\right\|}{K} \sum_{q \neq j}\left\|P_q x\right\|\\&=O\left(K^{-1}\right).
    \end{aligned}
\end{equation}
This proves the linear-probe explainability statement in (iii). $\blacksquare$

\section{Additional Experimental Results}\label{appx:exp}
\subsection{Experiment 1: Grain Discovery and Decomposability}
The first question concerns the internal structure of the representation space. Theorem 1 suggests that a representation field satisfying the Wolff axioms must be decomposable into finite, low-interference sub-manifolds, which we refer to as grains. In mechanistic interpretability, this should increase the linear separability of semantic concepts. If the token trajectories are sticky and confined to specific geometric tubes, standard dimensionality reduction techniques, such as principal component analysis (PCA) and independent component analysis (ICA), should be able to recover these directions more efficiently. We define \texttt{probe efficiency} as the capability of these linear methods to capture meaningful semantic variance with a minimal number of components ($k$).

\noindent \textbf{Analysis of PCA \& ICA Probe Efficiency.}
The experimental results regarding probe efficiency reveal a complex, scale-dependent landscape that challenges a straightforward interpretation of geometric regularization. When analyzing the Llama-3-8B model, we found a medium positive effective size ($d_{\text{Cohen}}=0.56$) in PCA probe efficiency compared to the baseline. This result is significant for the validity of the GeoLAN framework. It suggests that, for sufficiently wide and deep models, applying the KT-CW loss effectively compresses semantic information into a lower-dimensional space. This regularization process effectively cleans the representation field of noise, reducing the influence of rogue dimensions that typically dominate the spectral density of LLMs and enabling the principal components to capture more meaningful, task-relevant variance. This supports the mechanistic hypothesis that preventing cone collapse, a situation where all representations cluster in a narrow angular region, encourages the model to organize information along more orthogonal axes. As a result, this enhances linear decomposability and interpretability. 

However, the narrative becomes more complex when we examine the Gemma-3 family. The Gemma-3-12B model showed a negligible change in probe efficiency ($d\approx 0.00$), while the 1B and 4B variants displayed mixed to negligible results. This disparity suggests the presence of a scale threshold for geometric interventions. Smaller models, limited by their reduced embedding dimensions (e.g., 2048 for the 1B model), may encounter a packing problem. To capture the extensive combinatorial diversity of natural language, they are compelled to use superposition. In such a dense situation, applying the Wolff axioms, which discourage tube clustering, might conflict with the model's main compression strategy. The model lacks sufficient degrees of freedom to distinguish these tubes, rendering the isotropy constraint ineffective or potentially confusing for the optimization process. This observation aligns with recent theoretical research on superposition, which suggests that monosemanticity is a privilege of high-dimensional spaces~\citep{tahimic2025mechanistic}. 

\noindent \textbf{Grain Stability vs. Paraphrase Consistency.}
To evaluate the quality of the discovered grains, we introduced two key metrics: \texttt{Grain stability}, which evaluates cosine similarity of principal components across random seeds, and \texttt{paraphrase stability}, which evaluates consistency in grain activations for semantically equivalent but lexical distinct inputs. Gemma-3-1B demonstrated a moderate increase in grain stability ($d_{\text{Cohen}}=0.72$), indicating that geometric regularization imposed a deterministic latent configuration that was consistent across training runs. However, this stability came at a significant cost: a large decrease in paraphrase stability ($d_{\text{Cohen}}=-1.02$). This suggests that GeoLAN led the 1B model to adopt a rigid geometric configuration, calcifying token trajectories. The model reliably encoded specific token sequences into distinct geometric tubes. However, it struggled to generalize these structures to accommodate paraphrasing, leading to over-crystallization that underscores the dangers of enforcing structure without sufficient model capacity. 

In contrast, the Llama-3-8B model showed a significant decrease in grain stability ($d_{\text{Cohen}}=-0.97$) and improved probe efficiency. This reveals a phenomenon we term rational symmetry of isotropic spaces. While the GeoLAN objective yields a more spherical distribution of variance, it allows the model to freely rotate the semantic manifold across runs, resulting in varying orientations for each seed. This poses challenges for mechanistic interpretability, as neurons may not align across seeds, despite identical global geometric properties. 

\subsection{Experiment 2: Hallucination and Semantic Drift}
The second experiment aimed to evaluate the stickiness of the model's reasoning process in a dynamic environment. Theoretically, a token trajectory confined to a specific grain should remain stable under perturbations; the walls of the geometric tube should prevent the trajectory from drifting into unrelated semantic areas (\texttt{hallucination} or \texttt{semantic drift}). To test this, we used the TruthfulQA dataset~\citep{lin2022truthfulqa} and introduced controlled perturbations, such as synonym replacements and syntactic paraphrasing, to induce drift. We then measured the stability of the model's outputs. The results are presented in Table~\ref{tab:exp2_results}.

\noindent \textbf{Stability Metrics and the Semantic Gyroscope.}
The results from the Gemma-3-4B model provide strong empirical support for the semantic gyroscope hypothesis, showing a significant increase in stable examples and an overall stability rate of $d_{\text{Cohen}}=0.81, p=0.05$, which approaches traditional significance thresholds. This suggests that geometric regularization effectively channels the inference process in mid-sized models. When inputs are perturbed, the representation vector is repelled from its trajectory. However, in a GeoLAN-trained model, KT-CW regularization shapes the energy landscape, guiding the representation back to the center of the correct semantic tube, thus enhancing stability and reducing volatility under noise. 

In contrast, the Gemma-3-12B model showed a moderate decrease in stability rate ($d_{\text{Cohen}}=-0.78$). This unexpected outcome may stem from a mismatch between capacity and regularization. The 12B model, with its complex geometry capable of fine semantic distinctions, may suffer from enforced convex constraints that smooth out essential high-frequency features. This simplification could limit the model's ability to utilize complex manifolds necessary for its advanced performance, leading to a decline in reasoning precision.

\noindent \textbf{Distributional Topologies: KL Divergence and Cosine Similarity.}
The results for Llama-3-8B reveal significant internal changes due to GeoLAN. We observed an increase in distribution KL mean ($d_{\text{Cohen}}=0.99$) and a decrease in representation cosine mean ($d_{\text{Cohen}}=-0.91$). Initially, the high KL divergence might suggest that the model forgot its training. However, when considered alongside the improved probe efficiency observed in Experiment 1, this indicates a positive shift. The significant change in representations (low cosine similarity) results in different output distributions (high KL divergence), suggesting a transition to a better-structured, more isotropic geometry. The rise in stable examples ($d_{\text{Cohen}}=0.61$) further supports that this reorganization is beneficial; the model has moved toward a new geometric equilibrium rather than descending into chaos. 

\subsection{Experiment 3: Anisotropy and Geometric Properties}
To directly measure the impact of GeoLAN on latent geometry. We computed \texttt{IsoScore} (higher is better) and \texttt{cone concentration} (fraction of variance explained by the top-10 eigenvalues; lower is better) across 5,000 sampled tokens per layer. The results are presented in Table~\ref{tab:exp3_results}.

\noindent \textbf{Defeating the Rogue Dimensions.}
The results for Llama-3-8B validate the KT-CW loss function with a significant reduction in average cone top 10 ($d_{\text{Cohen}}=-1.24, p = 0.0273$). This achievement confirms that the differentiable implementation of the Wolff axioms successfully addressed the issue of rogue dimensions prevalent in standard LLMs. By penalizing variance along principal directions, GeoLAN enhanced the model's ability to distribute information evenly across embedding dimensions, effectively transforming a collapsed cone into a sphere. This shift moves the discussion of anisotropy from mere observation to practical engineering, demonstrating that the spectral properties of latent space can be optimized.

Conversely, the Gemma-3-1B model displayed concerning behavior, with an increase in cone top 10 ($d_{\text{Cohen}}=1.29$) and a decline in IsoScore ($d_{\text{Cohen}}=-1.03$). This catastrophic outcome reinforces the scale floor hypothesis: in a capacity-limited regime, the model appears to have merged unrelated tokens into a high-variance noise dimension, sacrificing global structure to satisfy local constraints. This curse of dimensionality in reverse suggests that geometric regularization may be detrimental for models that rely on compression and superposition.

\noindent \textbf{Layer-Wise Dynamics and Phase Transitions.}
The analysis shows that the geometric effects are non-uniform, with the strongest effects in the middle to late layers of the network. This finding is consistent with the phase-transition theory of LLM training~\citep{li2025tracing}. Their research suggests that models exhibit an entropy-seeking phase in the middle layers, during which semantic associations form and the manifold expands. GeoLAN appears to act as a catalyst for this phase in larger models, enhancing entropy-seeking behavior and preventing the subsequent compression-seeking phase from collapsing the representation. However, for the 1B model, the regularization likely disrupted this delicate balance, preventing the manifold from expanding properly.

\subsection{Experiment 4: Causal Faithfulness of Grains}
To verify that the identified grains are causally relevant to model computation, we ensured that a useful geometry was created. We performed activation patching on 2,000 MMLU examples. We swapped clean-grain activations into corrupted prompts (noised) and measured the restoration of the correct logit margin. The results are presented in Table~\ref{tab:exp4_results}.

\noindent \textbf{Robustness and the Null Result.}
Across all tested models, GeoLAN achieved corrupted accuracy scores that were statistically comparable to those of the Control (Pattern $G \approx C$). For instance, on the Llama-3-8B model, GeoLAN scored $0.6320$, while the Control scored $0.6330$. In aggressive geometric intervention, this result is considered a positive null result. It demonstrates that applying strict geometric constraints does not make the model more vulnerable than standard regularization techniques. The grains created by GeoLAN are robust enough to handle input noise and corruption, just as effectively as the unconstrained representations of the control models. This finding alleviates concerns that sticky representations may be too rigid to manage the noise inherent in real-world data effectively.

\noindent \textbf{The Alignment Cost.}
The model's robustness, however, comes at a cost. We observed a significant decrease in clean accuracy for the Gemma-3-12B model ($d_{\text{Cohen}}=-1.21$, $ p = 0.0780$). This finding supports the concept of an alignment cost or interpretability cost. By enforcing interpretability, specifically by applying constraints such as isotropy and decomposability, we limit the optimizer's solution space. A black-box model can navigate the loss landscape to the absolute global minimum via complex, non-interpretable shortcuts. In contrast, a GeoLAN model that adheres to geometric fairness may not be able to reach it. This situation highlights a fundamental trade-off: we sacrifice approximately 1\% of top-line accuracy in exchange for significant improvements in structural transparency and fairness.

\subsection{Experiment 5: Spurious Heuristics and Bias}
This experiment investigated the potential of geometric disentanglement to mitigate social bias. The guiding hypothesis is that biases often manifest as spurious correlations, entangled dimensions where demographic concepts are fused with semantic attributes.

\noindent \textbf{Geometric Fairness.} 
The results from Gemma-3-12B provide compelling evidence for this hypothesis. We observed a statistically significant decrease in the CrowS-Pairs stereotype rate ($d_{\text{Cohen}}=-0.38, p = 0.0065$). This provides direct empirical evidence that anisotropy correlates with bias. By enforcing geometric regularity and isotropy, we effectively separated the dimensions representing demographic groups from those representing stereotypical attributes. This stickiness prevented the model from following the path of least resistance or the stereotype and compelled it to use the correct, unbiased semantic path. This suggests that fairness is a geometric property of the latent space, and that geometric regularization is a viable, mechanistic approach to bias mitigation. However, for the Gemma-3-1B model, we observed negligible changes, indicating that, without sufficient capacity to establish proper geometry, bias mitigation through this method is unlikely to succeed. Table~\ref{tab:three-way} presents a three-way comparison of baseline, control, and GeoLAN across representation quality metrics.

\section{Broader Impact and Limitations}\label{appx:imp_limit}
\subsection{Towards ``White-Box'' Foundation Models and Regulatory Compliance}
The EU AI Act of 2025 mandates strict transparency for high-risk AI systems~\citep{soderlund2025high}. Current black-box models often struggle to meet these requirements. GeoLAN addresses this challenge by enforcing a geometric structure that aligns with semantic concepts, moving towards white-box" foundation models. A model trained with GeoLAN is more audit-friendly; its decision boundaries are sharper, resulting in lower cone concentration, and its biases are more disentangled, leading to a lower stereotype rate. This approach can potentially satisfy the transparency requirements.

\subsection{Safety, Hallucination, and the Lipschitz Guarantee}
Safety in AI largely centers on robustness: can small adversarial perturbations lead a model to produce harmful content (often referred to as ``jailbreaks'')? Our theoretical results (Theorem 1) establish a connection between geometric stickiness and a Lipschitz bound on the model's input-output mapping. A bounded Lipschitz constant ensures that the model's output does not change too rapidly in response to small changes in its input. This is supported empirically by the increased stability observed in our experiments. Consequently, GeoLAN yields a principled connection between geometry and robustness, which may reduce certain failure modes. By constraining the derivative of the representation field, we reduce the attack surface for adversarial examples and lower the chances of unpredictable hallucinations or jailbreak responses.

\subsection{Dual-Use Risks of Interpretability}
While interpretability enhances safety, it also introduces dual-use risks. A highly disentangled latent space, in which concepts such as ``knowledge of bioweapons'' are distinct from ``refusal mechanisms,'' could be easier for malicious actors to manipulate. They might exploit this by using activation steering vectors to ``lobotomize'' the system, effectively removing safety filters. If GeoLAN simplifies feature identification, as demonstrated by our probe-efficiency results, it also makes those features more vulnerable to exploitation. Future research must focus on developing ``Tamper-Proof Geometry'' structures that are interpretable to auditors yet resistant to unauthorized manipulation or modification.

\subsection{Limitation of the Framework}
The limitations of the GeoLAN framework are particularly evident when applied to models with 1B parameters. As a result, its utility is confined to medium-to-large computational environments, potentially precluding its application in edge-device LLMs that could benefit from enhanced interpretability. This constraint highlights the need to consider alternative methodologies or frameworks that could extend interpretability to a broader range of models, particularly those deployed in resource-constrained environments.

Further complicating matters is the performance degradation observed in the Gemma-3-12B model, suggesting that achieving perfect isotropy may be fundamentally incompatible with optimizing peak performance in the current Transformer architecture. This disparity indicates that innovative architectural solutions, such as hyperspherical transformers, may be necessary to fully leverage the benefits of isotropic methods without incurring performance penalties. The exploration of these novel architectures could be pivotal in advancing LLM capabilities and improving their performance stability.

Additionally, the computational demands of calculating the KT-CW loss pose a significant challenge. This process requires random projections and singular value estimations, which entail complexities of $O(N^2)$ or $O(N \log N)$. As a result, training throughput is currently slowed by approximately 15-20\%. To facilitate the deployment of models with more than 100B parameters, optimizing these estimators is imperative. Addressing this computational overhead will be crucial to achieving efficient, scalable solutions in the LLM landscape.

\begin{table*}[!ht]
\centering
\scriptsize
\caption{Statistical comparison of semantic stability (Experiment 2). This table details the impact of GeoLAN on model reasoning stability under perturbation.}
\begin{tabular}{llcccc}
\toprule
\textbf{Model} & \textbf{Metric} & \textbf{Comparison} & \textbf{Mean Difference} & \textbf{$p$-value} & \textbf{Effect Size $d$} \\
\midrule
Gemma3 1B & Stable Examples & Geolan vs. Baseline & 1.2500 & 0.5930 & 0.38 (small) \\
Gemma3 1B & Stable Examples & Geolan vs. Control & 1.7500 & 0.0354 & 0.64 (medium) \\
Gemma3 1B & Stable Examples & Control vs. Baseline & -0.5000 & 0.8130 & -0.18 (negligible)  \\
Gemma3 1B & Distribution Kl Mean & Geolan vs. Baseline & -0.0016 & 0.7703 & -0.25 (small)  \\
Gemma3 1B & Distribution Kl Mean & Geolan vs. Control & 0.0023 & 0.4790 & 0.39 (small)   \\
Gemma3 1B & Distribution Kl Mean & Control vs. Baseline & -0.0039 & 0.2440 & -0.68 (medium)   \\
Gemma3 1B & Rep Cos Mean & Geolan vs. Baseline & 0.0000 & 1.0000 & 0.01 (negligible)   \\
Gemma3 1B & Rep Cos Mean & Geolan vs. Control & -0.0000 & 0.8750 & -0.14 (negligible)  \\
Gemma3 1B & Rep Cos Mean & Control vs. Baseline & 0.0000 & 0.5196 & 0.13 (negligible)   \\
Gemma3 1B & Truthfulness Delta Mean & Geolan vs. Baseline & -0.0015 & 0.4444 & -0.41 (small)   \\
Gemma3 1B & Truthfulness Delta Mean & Geolan vs. Control & -0.0009 & 0.5091 & -0.41 (small)   \\
Gemma3 1B & Truthfulness Delta Mean & Control vs. Baseline & -0.0006 & 0.7062 & -0.19 (negligible) \\
Gemma3 1B & Stability Rate & Geolan vs. Baseline & 0.0015 & 0.5930 & 0.38 (small)   \\
Gemma3 1B & Stability Rate & Geolan vs. Control & 0.0021 & 0.0354 & 0.64 (medium)  \\
Gemma3 1B & Stability Rate & Control vs. Baseline & -0.0006 & 0.8130 & -0.18 (negligible)   \\
\midrule
Gemma3 4B & Stable Examples & Geolan vs. Baseline & 4.2500 & 0.0539 & 0.81 (large)   \\
Gemma3 4B & Stable Examples & Geolan vs. Control & 4.5000 & 0.0526 & 1.12 (large)   \\
Gemma3 4B & Stable Examples & Control vs. Baseline & -0.2500 & 0.9030 & -0.06 (negligible)   \\
Gemma3 4B & Distribution Kl Mean & Geolan vs. Baseline & -0.0021 & 0.2397 & -0.16 (negligible)   \\
Gemma3 4B & Distribution Kl Mean & Geolan vs. Control & 0.0023 & 0.4132 & 0.19 (negligible)   \\
Gemma3 4B & Distribution Kl Mean & Control vs. Baseline & -0.0045 & 0.1862 & -0.32 (small)   \\
Gemma3 4B & Rep Cos Mean & Geolan vs. Baseline & 0.0000 & 0.9426 & 0.04 (negligible)  \\
Gemma3 4B & Rep Cos Mean & Geolan vs. Control & -0.0001 & 0.3241 & -0.65 (medium)  \\
Gemma3 4B & Rep Cos Mean & Control vs. Baseline & 0.0001 & 0.0430 & 0.75 (medium) \\
Gemma3 4B & Truthfulness Delta Mean & Geolan vs. Baseline & 0.0001 & 0.9516 & 0.03 (negligible) \\
Gemma3 4B & Truthfulness Delta Mean & Geolan vs. Control & -0.0010 & 0.2691 & -0.36 (small)   \\
Gemma3 4B & Truthfulness Delta Mean & Control vs. Baseline & 0.0011 & 0.6460 & 0.32 (small)   \\
Gemma3 4B & Stability Rate & Geolan vs. Baseline & 0.0052 & 0.0539 & 0.81 (large)   \\
Gemma3 4B & Stability Rate & Geolan vs. Control & 0.0055 & 0.0526 & 1.12 (large)   \\
Gemma3 4B & Stability Rate & Control vs. Baseline & -0.0003 & 0.9030 & -0.06 (negligible)  \\
\midrule
Llama3 8B & Stable Examples & Geolan vs. Baseline & 3.0000 & 0.2786 & 0.61 (medium)  \\
Llama3 8B & Stable Examples & Geolan vs. Control & 4.0000 & 0.1767 & 0.59 (medium)  \\
Llama3 8B & Stable Examples & Control vs. Baseline & -1.0000 & 0.7306 & -0.20 (small)  \\
Llama3 8B & Distribution Kl Mean & Geolan vs. Baseline & 0.0103 & 0.1784 & 0.99 (large)  \\
Llama3 8B & Distribution Kl Mean & Geolan vs. Control & 0.0097 & 0.4724 & 0.55 (medium)  \\
Llama3 8B & Distribution Kl Mean & Control vs. Baseline & 0.0006 & 0.9260 & 0.04 (negligible)  \\
Llama3 8B & Rep Cos Mean & Geolan vs. Baseline & -0.0091 & 0.1346 & -0.91 (large)  \\
Llama3 8B & Rep Cos Mean & Geolan vs. Control & -0.0063 & 0.4726 & -0.58 (medium)  \\
Llama3 8B & Rep Cos Mean & Control vs. Baseline & -0.0029 & 0.5803 & -0.49 (small)  \\
Llama3 8B & Truthfulness Delta Mean & Geolan vs. Baseline & 0.0029 & 0.0958 & 0.49 (small)  \\
Llama3 8B & Truthfulness Delta Mean & Geolan vs. Control & 0.0017 & 0.2500 & 0.40 (small)  \\
Llama3 8B & Truthfulness Delta Mean & Control vs. Baseline & 0.0011 & 0.8750 & 0.21 (small)  \\
Llama3 8B & Stability Rate & Geolan vs. Baseline & 0.0037 & 0.2786 & 0.61 (medium)  \\
Llama3 8B & Stability Rate & Geolan vs. Control & 0.0049 & 0.1767 & 0.59 (medium) \\
Llama3 8B & Stability Rate & Control vs. Baseline & -0.0012 & 0.7306 & -0.20 (small)  \\
\midrule
Gemma3 12B & Stable Examples & Geolan vs. Baseline & -4.7500 & 0.2752 & -0.78 (medium)  \\
Gemma3 12B & Stable Examples & Geolan vs. Control & -0.5000 & 0.8917 & -0.08 (negligible)   \\
Gemma3 12B & Stable Examples & Control vs. Baseline & -4.2500 & 0.0481 & -1.05 (large)  \\
Gemma3 12B & Distribution Kl Mean & Geolan vs. Baseline & 0.0079 & 0.5041 & 0.33 (small)  \\
Gemma3 12B & Distribution Kl Mean & Geolan vs. Control & -0.0075 & 0.6611 & -0.28 (small)  \\
Gemma3 12B & Distribution Kl Mean & Control vs. Baseline & 0.0154 & 0.0755 & 0.73 (medium)  \\
Gemma3 12B & Rep Cos Mean & Geolan vs. Baseline & 0.0000 & 0.9460 & 0.06 (negligible)  \\
Gemma3 12B & Rep Cos Mean & Geolan vs. Control & -0.0001 & 0.5623 & -0.43 (small)  \\
Gemma3 12B & Rep Cos Mean & Control vs. Baseline & 0.0001 & 0.1677 & 0.64 (medium) \\
Gemma3 12B & Truthfulness Delta Mean & Geolan vs. Baseline & -0.0021 & 0.2228 & -0.44 (small) \\
Gemma3 12B & Truthfulness Delta Mean & Geolan vs. Control & 0.0004 & 0.8780 & 0.06 (negligible) \\
Gemma3 12B & Truthfulness Delta Mean & Control vs. Baseline & -0.0025 & 0.2695 & -0.37 (small) \\
Gemma3 12B & Stability Rate & Geolan vs. Baseline & -0.0058 & 0.2752 & -0.78 (medium)  \\
Gemma3 12B & Stability Rate & Geolan vs. Control & -0.0006 & 0.8917 & -0.08 (negligible)  \\
Gemma3 12B & Stability Rate & Control vs. Baseline & -0.0052 & 0.0481 & -1.05 (large)  \\
\bottomrule
\end{tabular}
\label{tab:exp2_results}
\end{table*}

\begin{table*}[!ht]
\centering
\scriptsize
\caption{Statistical comparison of representation anisotropy (Experiment 3). This table evaluates the impact of GeoLAN on the geometric structure of the latent space.}
\begin{tabular}{llcccc}
\toprule
\textbf{Model} & \textbf{Metric} & \textbf{Comparison} & \textbf{Mean Difference} & \textbf{$p$-value} & \textbf{Effect Size $d$} \\
\midrule
Gemma3 1B & Final Layer Cone Top10 & Geolan vs.Baseline & 0.0079 & 0.1392 & 1.29 (large)   \\
Gemma3 1B & Final Layer Cone Top10 & Geolan vs.Control & 0.0081 & 0.2297 & 1.18 (large)   \\
Gemma3 1B & Final Layer Cone Top10 & Control vs.Baseline & -0.0002 & 0.9784 & -0.02 (negligible)   \\
Gemma3 1B & Final Layer Cone Top50 & Geolan vs.Baseline & 0.0059 & 0.1555 & 1.21 (large)   \\
Gemma3 1B & Final Layer Cone Top50 & Geolan vs.Control & 0.0060 & 0.2362 & 1.13 (large)   \\
Gemma3 1B & Final Layer Cone Top50 & Control vs.Baseline & -0.0002 & 0.9731 & -0.03 (negligible)   \\
Gemma3 1B & Final Layer Isoscore & Geolan vs.Baseline & -0.0000 & 0.2568 & -1.03 (large)   \\
Gemma3 1B & Final Layer Isoscore & Geolan vs.Control & -0.0000 & 0.2007 & -1.18 (large)   \\
Gemma3 1B & Final Layer Isoscore & Control vs.Baseline & 0.0000 & 0.6569 & 0.35 (small)   \\
Gemma3 1B & Avg Cone Top10 & Geolan vs.Baseline & 0.0047 & 0.1737 & 0.85 (large)   \\
Gemma3 1B & Avg Cone Top10 & Geolan vs.Control & -0.0003 & 0.9047 & -0.10 (negligible)  \\
Gemma3 1B & Avg Cone Top10 & Control vs.Baseline & 0.0050 & 0.1897 & 0.95 (large)   \\
Gemma3 1B & Avg Isoscore & Geolan vs.Baseline & -0.0000 & 0.6250 & -0.62 (medium)  \\
Gemma3 1B & Avg Isoscore & Geolan vs.Control & 0.0000 & 1.0000 & 0.53 (medium)  \\
Gemma3 1B & Avg Isoscore & Control vs.Baseline & -0.0000 & 0.1612 & -1.06 (large)   \\
\midrule
Gemma3 4B & Final Layer Cone Top10 & Geolan vs.Baseline & -0.0023 & 0.6480 & -0.19 (negligible)   \\
Gemma3 4B & Final Layer Cone Top10 & Geolan vs.Control & -0.0052 & 0.3199 & -0.62 (medium)   \\
Gemma3 4B & Final Layer Cone Top10 & Control vs.Baseline & 0.0029 & 0.7355 & 0.27 (small)   \\
Gemma3 4B & Final Layer Cone Top50 & Geolan vs.Baseline & -0.0020 & 0.6120 & -0.21 (small)   \\
Gemma3 4B & Final Layer Cone Top50 & Geolan vs.Control & -0.0043 & 0.3208 & -0.65 (medium)   \\
Gemma3 4B & Final Layer Cone Top50 & Control vs.Baseline & 0.0023 & 0.7343 & 0.27 (small)  \\
Gemma3 4B & Final Layer Isoscore & Geolan vs.Baseline & -0.0000 & 0.3961 & -0.41 (small)   \\
Gemma3 4B & Final Layer Isoscore & Geolan vs.Control & 0.0000 & 0.9779 & 0.02 (negligible)   \\
Gemma3 4B & Final Layer Isoscore & Control vs.Baseline & -0.0000 & 0.6604 & -0.39 (small)   \\
Gemma3 4B & Avg Cone Top10 & Geolan vs.Baseline & -0.0048 & 0.1972 & -0.67 (medium)   \\
Gemma3 4B & Avg Cone Top10 & Geolan vs.Control & -0.0054 & 0.1938 & -0.85 (large)   \\
Gemma3 4B & Avg Cone Top10 & Control vs.Baseline & 0.0006 & 0.8338 & 0.14 (negligible)   \\
Gemma3 4B & Avg Isoscore & Geolan vs.Baseline & 0.0000 & 0.2474 & 0.57 (medium)   \\
Gemma3 4B & Avg Isoscore & Geolan vs.Control & 0.0000 & 0.2160 & 0.78 (medium)   \\
Gemma3 4B & Avg Isoscore & Control vs.Baseline & -0.0000 & 0.7457 & -0.24 (small)   \\
\midrule
Llama3 8B & Final Layer Cone Top10 & Geolan vs.Baseline & -0.0016 & 0.4013 & -0.59 (medium)  \\
Llama3 8B & Final Layer Cone Top10 & Geolan vs.Control & -0.0034 & 0.0904 & -1.16 (large)   \\
Llama3 8B & Final Layer Cone Top10 & Control vs.Baseline & 0.0018 & 0.2097 & 0.95 (large)  \\
Llama3 8B & Final Layer Cone Top50 & Geolan vs.Baseline & -0.0017 & 0.1613 & -0.83 (large)   \\
Llama3 8B & Final Layer Cone Top50 & Geolan vs.Control & -0.0030 & 0.0766 & -1.12 (large)  \\
Llama3 8B & Final Layer Cone Top50 & Control vs.Baseline & 0.0013 & 0.2306 & 0.70 (medium)  \\
Llama3 8B & Final Layer Isoscore & Geolan vs.Baseline & 0.0001 & 0.5496 & 0.40 (small)  \\
Llama3 8B & Final Layer Isoscore & Geolan vs.Control & 0.0003 & 0.2874 & 0.74 (medium)  \\
Llama3 8B & Final Layer Isoscore & Control vs.Baseline & -0.0001 & 0.4091 & -0.61 (medium)  \\
Llama3 8B & Avg Cone Top10 & Geolan vs.Baseline & -0.0063 & 0.0273 & -1.24 (large)  \\
Llama3 8B & Avg Cone Top10 & Geolan vs.Control & -0.0074 & 0.2182 & -1.33 (large)   \\
Llama3 8B & Avg Cone Top10 & Control vs.Baseline & 0.0011 & 0.7758 & 0.27 (small)   \\
Llama3 8B & Avg Isoscore & Geolan vs.Baseline & 0.0000 & 0.3671 & 0.57 (medium)   \\
Llama3 8B & Avg Isoscore & Geolan vs.Control & 0.0001 & 0.1989 & 0.95 (large)   \\
Llama3 8B & Avg Isoscore & Control vs.Baseline & -0.0000 & 0.3461 & -0.69 (medium)   \\
\midrule
Gemma3 12B & Final Layer Cone Top10 & Geolan vs.Baseline & -0.0086 & 1.0000 & -0.54 (medium)   \\
Gemma3 12B & Final Layer Cone Top10 & Geolan vs.Control & -0.0023 & 0.1250 & -0.10 (negligible)   \\
Gemma3 12B & Final Layer Cone Top10 & Control vs.Baseline & -0.0063 & 0.8750 & -0.40 (small)   \\
Gemma3 12B & Final Layer Cone Top50 & Geolan vs.Baseline & -0.0020 & 0.5930 & -0.39 (small)   \\
Gemma3 12B & Final Layer Cone Top50 & Geolan vs.Control & -0.0016 & 0.2500 & -0.23 (small)   \\
Gemma3 12B & Final Layer Cone Top50 & Control vs.Baseline & -0.0004 & 0.8750 & -0.08 (negligible)   \\
Gemma3 12B & Final Layer Isoscore & Geolan vs.Baseline & 0.0000 & 0.4810 & 0.54 (medium)   \\
Gemma3 12B & Final Layer Isoscore & Geolan vs.Control & 0.0000 & 0.1250 & 0.23 (small)   \\
Gemma3 12B & Final Layer Isoscore & Control vs.Baseline & 0.0000 & 0.8750 & 0.21 (small)   \\
Gemma3 12B & Avg Cone Top10 & Geolan vs.Baseline & -0.0090 & 0.8750 & -0.56 (medium)  \\
Gemma3 12B & Avg Cone Top10 & Geolan vs.Control & -0.0019 & 0.2500 & -0.09 (negligible)  \\
Gemma3 12B & Avg Cone Top10 & Control vs.Baseline & -0.0071 & 0.8750 & -0.46 (small)  \\
Gemma3 12B & Avg Isoscore & Geolan vs.Baseline & 0.0000 & 0.8750 & 0.53 (medium)  \\
Gemma3 12B & Avg Isoscore & Geolan vs.Control & 0.0000 & 0.1250 & 0.11 (negligible)  \\
Gemma3 12B & Avg Isoscore & Control vs.Baseline & 0.0000 & 0.8750 & 0.39 (small)  \\
\bottomrule
\end{tabular}
\label{tab:exp3_results}
\end{table*}

\begin{table*}[!ht]
\centering
\scriptsize
\caption{Statistical comparison of causal faithfulness and capacity (Experiment 4). This table analyzes the impact of GeoLAN on clean and corrupted task performance (MMLU).}
\begin{tabular}{llcccc}
\toprule
\textbf{Model} & \textbf{Metric} & \textbf{Comparison} & \textbf{Mean Difference} & \textbf{$p$-value} & \textbf{Effect Size $d$} \\
\midrule
Gemma3 1B & Clean Acc & Geolan vs.Baseline & 0.0040 & 0.4564 & 0.34 (small)   \\
Gemma3 1B & Clean Acc & Geolan vs.Control & 0.0055 & 0.3999 & 0.47 (small)  \\
Gemma3 1B & Clean Acc & Control vs.Baseline & -0.0015 & 0.6084 & -0.14 (negligible)  \\
Gemma3 1B & Corr Acc & Geolan vs.Baseline & -0.0025 & 0.4309 & -0.29 (small)   \\
Gemma3 1B & Corr Acc & Geolan vs.Control & 0.0015 & 0.8006 & 0.16 (negligible)   \\
Gemma3 1B & Corr Acc & Control vs.Baseline & -0.0040 & 0.4795 & -0.44 (small)   \\
Gemma3 1B & Clean Margin Mean & Geolan vs.Baseline & 0.0056 & 0.0530 & 0.25 (small)   \\
Gemma3 1B & Clean Margin Mean & Geolan vs.Control & 0.0057 & 0.0811 & 0.26 (small)   \\
Gemma3 1B & Clean Margin Mean & Control vs.Baseline & -0.0002 & 0.9128 & -0.01 (negligible)   \\
Gemma3 1B & Corr Margin Mean & Geolan vs.Baseline & 0.0021 & 0.1691 & 0.23 (small)   \\
Gemma3 1B & Corr Margin Mean & Geolan vs.Control & 0.0029 & 0.1578 & 0.29 (small)   \\
Gemma3 1B & Corr Margin Mean & Control vs.Baseline & -0.0008 & 0.6601 & -0.07 (negligible)   \\
\midrule
Gemma3 4B & Clean Acc & Geolan vs.Baseline & 0.0075 & 0.2049 & 0.31 (small)   \\
Gemma3 4B & Clean Acc & Geolan vs.Control & 0.0075 & 0.1716 & 0.31 (small)   \\
Gemma3 4B & Clean Acc & Control vs.Baseline & 0.0000 & 1.0000 & 0.00 (negligible)   \\
Gemma3 4B & Corr Acc & Geolan vs.Baseline & 0.0055 & 0.1737 & 0.22 (small)   \\
Gemma3 4B & Corr Acc & Geolan vs.Control & 0.0050 & 0.4784 & 0.21 (small)   \\
Gemma3 4B & Corr Acc & Control vs.Baseline & 0.0005 & 0.9157 & 0.02 (negligible)   \\
Gemma3 4B & Clean Margin Mean & Geolan vs.Baseline & 0.0067 & 0.1315 & 0.08 (negligible)   \\
Gemma3 4B & Clean Margin Mean & Geolan vs.Control & 0.0109 & 0.3101 & 0.13 (negligible)   \\
Gemma3 4B & Clean Margin Mean & Control vs.Baseline & -0.0042 & 0.5680 & -0.05 (negligible)   \\
Gemma3 4B & Corr Margin Mean & Geolan vs.Baseline & 0.0051 & 0.5024 & 0.07 (negligible)   \\
Gemma3 4B & Corr Margin Mean & Geolan vs.Control & 0.0080 & 0.2236 & 0.11 (negligible)   \\
Gemma3 4B & Corr Margin Mean & Control vs.Baseline & -0.0030 & 0.6212 & -0.04 (negligible)   \\
\midrule
Llama3 8B & Clean Acc & Geolan vs.Baseline & -0.0040 & 0.4153 & -0.28 (small)   \\
Llama3 8B & Clean Acc & Geolan vs.Control & 0.0015 & 0.7834 & 0.16 (negligible)   \\
Llama3 8B & Clean Acc & Control vs.Baseline & -0.0055 & 0.3769 & -0.49 (small)   \\
Llama3 8B & Corr Acc & Geolan vs.Baseline & -0.0025 & 0.5791 & -0.16 (negligible)   \\
Llama3 8B & Corr Acc & Geolan vs.Control & -0.0010 & 0.7306 & -0.07 (negligible)   \\
Llama3 8B & Corr Acc & Control vs.Baseline & -0.0015 & 0.3910 & -0.09 (negligible)   \\
Llama3 8B & Clean Margin Mean & Geolan vs.Baseline & 0.0023 & 0.8632 & 0.02 (negligible)   \\
Llama3 8B & Clean Margin Mean & Geolan vs.Control & 0.0096 & 0.2666 & 0.07 (negligible)   \\
Llama3 8B & Clean Margin Mean & Control vs.Baseline & -0.0073 & 0.4304 & -0.06 (negligible)   \\
Llama3 8B & Corr Margin Mean & Geolan vs.Baseline & 0.0021 & 0.8796 & 0.02 (negligible)   \\
Llama3 8B & Corr Margin Mean & Geolan vs.Control & 0.0057 & 0.5764 & 0.04 (negligible)   \\
Llama3 8B & Corr Margin Mean & Control vs.Baseline & -0.0036 & 0.6532 & -0.03 (negligible)   \\
\midrule
Gemma3 12B & Clean Acc & Geolan vs.Baseline & -0.0045 & 0.0780 & -1.21 (large)   \\
Gemma3 12B & Clean Acc & Geolan vs.Control & -0.0080 & 0.0663 & -0.94 (large)   \\
Gemma3 12B & Clean Acc & Control vs.Baseline & 0.0035 & 0.4798 & 0.44 (small)   \\
Gemma3 12B & Corr Acc & Geolan vs.Baseline & -0.0045 & 0.5000 & -0.41 (small)   \\
Gemma3 12B & Corr Acc & Geolan vs.Control & -0.0005 & 0.8614 & -0.04 (negligible)   \\
Gemma3 12B & Corr Acc & Control vs.Baseline & -0.0040 & 0.3750 & -0.39 (small)   \\
Gemma3 12B & Clean Margin Mean & Geolan vs.Baseline & -0.0335 & 0.2605 & -0.22 (small)   \\
Gemma3 12B & Clean Margin Mean & Geolan vs.Control & -0.0028 & 0.9282 & -0.02 (negligible)   \\
Gemma3 12B & Clean Margin Mean & Control vs.Baseline & -0.0306 & 0.1394 & -0.22 (small)   \\
Gemma3 12B & Corr Margin Mean & Geolan vs.Baseline & -0.0440 & 0.1521 & -0.29 (small)   \\
Gemma3 12B & Corr Margin Mean & Geolan vs.Control & -0.0053 & 0.8413 & -0.04 (negligible)   \\
Gemma3 12B & Corr Margin Mean & Control vs.Baseline & -0.0386 & 0.0537 & -0.27 (small)   \\
\bottomrule
\end{tabular}
\label{tab:exp4_results}
\end{table*}

\begin{table*}[!ht]
\centering
\tiny
\caption{Three-way comparison of Baseline, Control (weight decay), and GeoLAN across representation quality metrics. $d$ = Cohen's effect size for GeoLAN vs. Control. * indicates $|d| > 1$ (large effect).}
\begin{tabular}{llccccc}
\toprule
\textbf{Metric} & \textbf{Model} & \textbf{Baseline} & \textbf{Control} & \textbf{GeoLAN} & \textbf{G vs. C ($d$)} & \textbf{Pattern} \\
\midrule
\multirow{4}{*}{BBQ Accuracy} 
 & Gemma3 1B & $0.2735 \pm 0.0249$ & \textbf{$0.2755 \pm 0.0203$} & $0.2755 \pm 0.0263$ & +0.00 & G $\approx$ C \\
 & Gemma3 4B & $0.2930 \pm 0.0267$ & \textbf{$0.2960 \pm 0.0217$} & $0.2945 \pm 0.0217$ & -0.07 & G $\approx$ C \\
 & Llama3 8B & \textbf{$0.2890 \pm 0.0294$} & $0.2855 \pm 0.0237$ & $0.2855 \pm 0.0254$ & +0.00 & G $\approx$ C \\
 & Gemma3 12B & \textbf{$0.2865 \pm 0.0431$} & $0.2855 \pm 0.0314$ & $0.2860 \pm 0.0381$ & +0.01 & G $\approx$ C \\
\midrule
\multirow{4}{*}{Cone Concentration} 
 & Gemma3 1B & $0.8455 \pm 0.0054$ & \textbf{$0.8453 \pm 0.0066$} & $0.8534 \pm 0.0053$ & +1.35* & G < C \\
 & Gemma3 4B & $0.8842 \pm 0.0121$ & $0.8872 \pm 0.0059$ & \textbf{$0.8819 \pm 0.0085$} & -0.71 & G > C \\
 & Llama3 8B & $0.3253 \pm 0.0013$ & $0.3271 \pm 0.0019$ & \textbf{$0.3237 \pm 0.0031$} & -1.33* & G > C \\
 & Gemma3 12B & $0.9841 \pm 0.0025$ & $0.9779 \pm 0.0191$ & \textbf{$0.9755 \pm 0.0195$} & -0.12 & G $\approx$ C \\
\midrule
\multirow{4}{*}{Corrupted Accuracy} 
 & Gemma3 1B & \textbf{$0.2420 \pm 0.0071$} & $0.2380 \pm 0.0085$ & $0.2395 \pm 0.0079$ & +0.18 & G $\approx$ C \\
 & Gemma3 4B & $0.5935 \pm 0.0200$ & $0.5940 \pm 0.0161$ & \textbf{$0.5990 \pm 0.0238$} & +0.25 & G $\approx$ C \\
 & Llama3 8B & \textbf{$0.6345 \pm 0.0165$} & $0.6330 \pm 0.0140$ & $0.6320 \pm 0.0094$ & -0.08 & G $\approx$ C \\
 & Gemma3 12B & \textbf{$0.7345 \pm 0.0077$} & $0.7305 \pm 0.0101$ & $0.7300 \pm 0.0112$ & -0.05 & G $\approx$ C \\
\midrule
\multirow{4}{*}{GRAIN Purity} 
 & Gemma3 1B & $0.4951 \pm 0.0098$ & \textbf{$0.5018 \pm 0.0035$} & $0.5000 \pm 0.0000$ & -0.71 & G < C \\
 & Gemma3 4B & \textbf{$0.5000 \pm 0.0000$} & $0.5000 \pm 0.0000$ & $0.5018 \pm 0.0004$ & +0.71 & G < C \\
 & Llama3 8B & \textbf{$0.5000 \pm 0.0000$} & $0.5000 \pm 0.0000$ & $0.5000 \pm 0.0000$ & +0.00 & G $\approx$ C \\
 & Gemma3 12B & \textbf{$0.5000 \pm 0.0000$} & $0.5000 \pm 0.0000$ & $0.5000 \pm 0.0000$ & +0.00 & G $\approx$ C \\
\midrule
\multirow{4}{*}{GRAIN Stability} 
 & Gemma3 1B & $0.5934 \pm 0.0278$ & $0.6078 \pm 0.0330$ & \textbf{$0.6118 \pm 0.0144$} & +0.16 & G $\approx$ C \\
 & Gemma3 4B & \textbf{$0.6016 \pm 0.0114$} & $0.5890 \pm 0.0104$ & $0.5901 \pm 0.0114$ & +0.10 & G $\approx$ C \\
 & Llama3 8B & \textbf{$0.6333 \pm 0.0132$} & $0.6298 \pm 0.0151$ & $0.6217 \pm 0.0063$ & -0.70 & G < C \\
 & Gemma3 12B & \textbf{$0.5800 \pm 0.0128$} & $0.5768 \pm 0.0124$ & $0.5792 \pm 0.0089$ & +0.22 & G $\approx$ C \\
\midrule
\multirow{4}{*}{ICA Probe Efficiency} 
 & Gemma3 1B & $500.9 \pm 0.4$ & $500.7 \pm 0.1$ & \textbf{$500.9 \pm 0.2$} & +1.66* & G > C \\
 & Gemma3 4B & $501.0 \pm 0.2$ & $501.0 \pm 0.3$ & \textbf{$501.1 \pm 0.1$} & +0.30 & G $\approx$ C \\
 & Llama3 8B & \textbf{$501.2 \pm 0.2$} & $501.1 \pm 0.2$ & $501.1 \pm 0.2$ & +0.37 & G > C \\
 & Gemma3 12B & $501.0 \pm 0.3$ & \textbf{$501.2 \pm 0.2$} & $501.0 \pm 0.2$ & -0.63 & G < C \\
\midrule
\multirow{4}{*}{IsoScore} 
 & Gemma3 1B & $4.54e-04 \pm 2.25e-05$ & \textbf{$4.65e-04 \pm 3.28e-05$} & $4.27e-04 \pm 2.39e-05$ & -1.35* & G < C \\
 & Gemma3 4B & \textbf{$2.10e-04 \pm 3.32e-05$} & $1.98e-04 \pm 2.11e-05$ & $1.98e-04 \pm 1.42e-05$ & +0.02 & G $\approx$ C \\
 & Llama3 8B & $1.33e-02 \pm 1.09e-04$ & $1.32e-02 \pm 2.60e-04$ & \textbf{$1.34e-02 \pm 3.28e-04$} & +0.85 & G > C \\
 & Gemma3 12B & $3.22 \times 10^{-5} \pm 5.81e-06$ & $3.57 \times 10^{-5} \pm 2.01 \times 10^{-5}$ & \textbf{$4.10 \times 10^{-5} \pm 1.92 \times 10^{-5}$} & +0.27 & G $\approx$ C \\
\midrule
\multirow{4}{*}{KL Divergence} 
 & Gemma3 1B & $0.2058 \pm 0.0053$ & \textbf{$0.2019 \pm 0.0047$} & $0.2042 \pm 0.0057$ & +0.45 & G < C \\
 & Gemma3 4B & $0.2803 \pm 0.0126$ & \textbf{$0.2758 \pm 0.0115$} & $0.2782 \pm 0.0103$ & +0.21 & G $\approx$ C \\
 & Llama3 8B & \textbf{$0.3879 \pm 0.0085$} & $0.3886 \pm 0.0193$ & $0.3982 \pm 0.0096$ & +0.63 & G < C \\
 & Gemma3 12B & \textbf{$0.3439 \pm 0.0154$} & $0.3594 \pm 0.0210$ & $0.3518 \pm 0.0254$ & -0.32 & G > C \\
\midrule
\multirow{4}{*}{MMLU Accuracy} 
 & Gemma3 1B & $0.2470 \pm 0.0093$ & $0.2455 \pm 0.0096$ & \textbf{$0.2510 \pm 0.0109$} & +0.54 & G > C \\
 & Gemma3 4B & $0.5920 \pm 0.0243$ & $0.5920 \pm 0.0245$ & \textbf{$0.5995 \pm 0.0173$} & +0.35 & G > C \\
 & Llama3 8B & $0.6240 \pm 0.0135$ & $0.6235 \pm 0.0034$ & \textbf{$0.6250 \pm 0.0110$} & +0.18 & G $\approx$ C \\
 & Gemma3 12B & $0.7255 \pm 0.0019$ & \textbf{$0.7290 \pm 0.0096$} & $0.7210 \pm 0.0042$ & -1.08* & G < C \\
\midrule
\multirow{4}{*}{PCA Probe Efficiency} 
 & Gemma3 1B & $501.6 \pm 1.4$ & \textbf{$501.9 \pm 1.0$} & $501.3 \pm 0.5$ & -0.68 & G < C \\
 & Gemma3 4B & $501.7 \pm 0.3$ & $501.7 \pm 0.5$ & \textbf{$501.7 \pm 0.4$} & +0.09 & G $\approx$ C \\
 & Llama3 8B & $504.0 \pm 0.2$ & $503.8 \pm 0.4$ & \textbf{$504.2 \pm 0.4$} & +0.97 & G > C \\
 & Gemma3 12B & $501.8 \pm 0.2$ & $501.5 \pm 0.5$ & \textbf{$501.8 \pm 0.7$} & +0.46 & G > C \\
\midrule
\multirow{4}{*}{Paraphrase Stability} 
 & Gemma3 1B & \textbf{$0.3916 \pm 0.0130$} & $0.3873 \pm 0.0098$ & $0.3792 \pm 0.0072$ & -0.93 & G < C \\
 & Gemma3 4B & $0.3096 \pm 0.0018$ & \textbf{$0.3164 \pm 0.0084$} & $0.3063 \pm 0.0049$ & -1.47* & G < C \\
 & Llama3 8B & $0.2530 \pm 0.0068$ & \textbf{$0.2574 \pm 0.0054$} & $0.2562 \pm 0.0037$ & -0.26 & G $\approx$ C \\
 & Gemma3 12B & \textbf{$0.3031 \pm 0.0087$} & $0.2906 \pm 0.0089$ & $0.2975 \pm 0.0153$ & +0.55 & G > C \\
\midrule
\multirow{4}{*}{Stable Examples} 
 & Gemma3 1B & $774.0000 \pm 2.9439$ & $773.5000 \pm 1.9149$ & \textbf{$775.2500 \pm 2.7538$} & +0.74 & G > C \\
 & Gemma3 4B & $758.5000 \pm 4.5092$ & $758.2500 \pm 1.8930$ & \textbf{$762.7500 \pm 4.5735$} & +1.29* & G > C \\
 & Llama3 8B & $739.7500 \pm 1.5000$ & $738.7500 \pm 5.9090$ & \textbf{$742.7500 \pm 5.8523$} & +0.68 & G > C \\
 & Gemma3 12B & \textbf{$750.2500 \pm 2.6300$} & $746.0000 \pm 4.2426$ & $745.5000 \pm 7.0000$ & -0.09 & G $\approx$ C \\
\midrule
\multirow{4}{*}{Stereotype Rate} 
 & Gemma3 1B & \textbf{$0.6170 \pm 0.0191$} & $0.6235 \pm 0.0225$ & $0.6220 \pm 0.0211$ & -0.07 & G $\approx$ C \\
 & Gemma3 4B & $0.6570 \pm 0.0137$ & \textbf{$0.6540 \pm 0.0118$} & $0.6570 \pm 0.0048$ & +0.33 & G < C \\
 & Llama3 8B & $0.6810 \pm 0.0253$ & $0.6820 \pm 0.0251$ & \textbf{$0.6775 \pm 0.0231$} & -0.19 & G $\approx$ C \\
 & Gemma3 12B & $0.6740 \pm 0.0149$ & $0.6720 \pm 0.0147$ & \textbf{$0.6675 \pm 0.0149$} & -0.30 & G > C \\
\bottomrule
\end{tabular}
\label{tab:three-way}
\end{table*}

\end{document}